\pdfoutput=1

\documentclass[11pt]{article}

\usepackage{acl}
\usepackage[title]{appendix}

\usepackage{times}
\usepackage{latexsym}
\usepackage{graphicx}
\usepackage{algorithm}
\usepackage{algpseudocode}
\usepackage[T1]{fontenc}

\usepackage[utf8]{inputenc}

\usepackage{inconsolata}
\usepackage[utf8]{inputenc}
\usepackage{microtype}
\usepackage{todonotes}
\makeatletter
\newcommand*\iftodonotes{\if@todonotes@disabled\expandafter\@secondoftwo\else\expandafter\@firstoftwo\fi}  
\makeatother

\title{What do tokens know about their characters and how do they know it?}

\author{ Ayush Kaushal\\ \texttt{\href{mailto:ayush}{ayushk4@utexas.edu}}\\The University of Texas at Austin \And Kyle Mahowald\\ \texttt{\href{mailto:kyle}{mahowald@utexas.edu}}\\The University of Texas at Austin
}

\begin{document}
\maketitle

\begin{abstract}

Pre-trained language models (PLMs) that use subword tokenization schemes can succeed at a variety of language tasks that require character-level information, despite lacking explicit access to the character composition of tokens.
Here, studying a range of models (e.g., GPT-J, BERT, RoBERTa, GloVe), we probe what word pieces encode about character-level information by training classifiers to predict the presence or absence of a particular alphabetical character in a token, based on its embedding (e.g., probing whether the model embedding for "cat" encodes that it contains the character "a").
We find that these models robustly encode character-level information and, in general, larger models perform better at the task.
We show that these results generalize to characters from non-Latin alphabets (Arabic, Devanagari, and Cyrillic).
Then, through a series of experiments and analyses, we investigate the mechanisms through which PLMs acquire English-language character information during training and argue that this knowledge is acquired through multiple phenomena, including a systematic relationship between particular characters and particular parts of speech, as well as natural variability in the tokenization of related strings.
\end{abstract}

\section{Introduction and Motivation}

\begin{figure}[t]
    \centering
    \includegraphics[width=0.9\columnwidth]{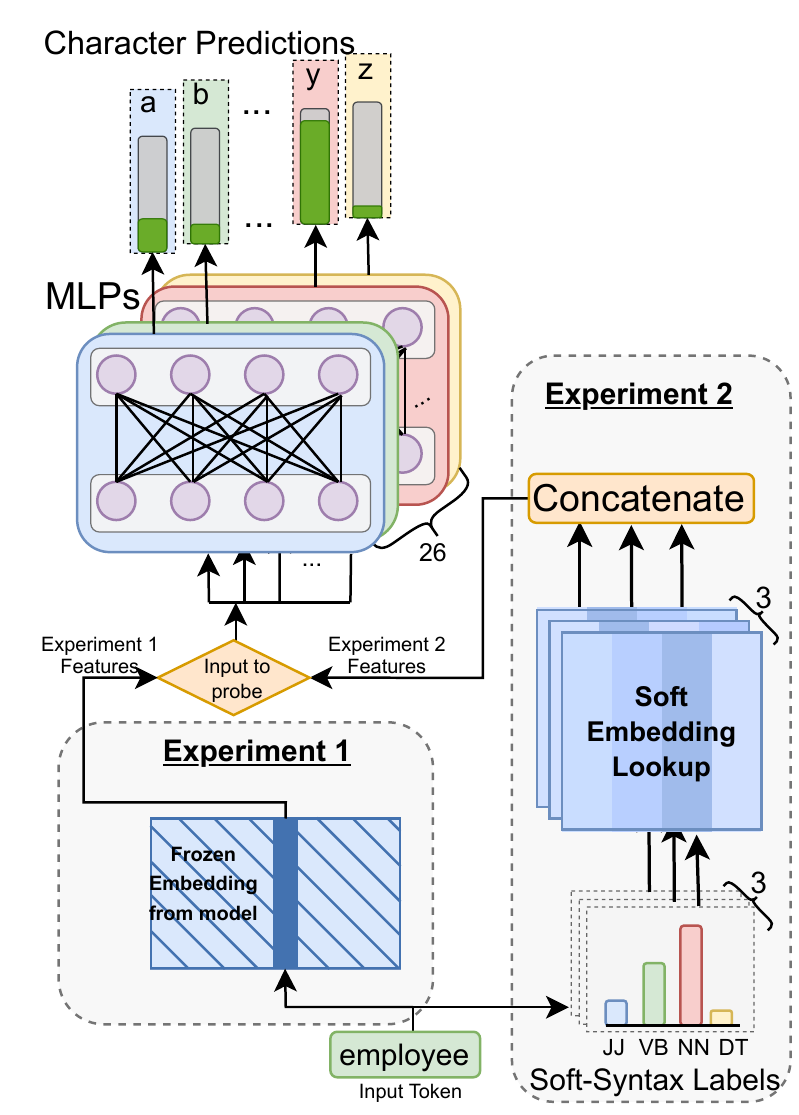}
    \caption{Overview of our probing setup. In Experiment 1, the input is a model embedding and we train MLPs to classify whether a particular character (e.g., "a") occurs in a particular token (e.g, "employee"). In Experiment 2, we use syntactic features as input, rather than model embeddings, to train our probe.}
    \label{fig:expt_1_diagram}
\end{figure}

The dominant class of models in NLP \citep[pre-trained transformer models; ][]{brown2020language,devlin-etal-2019-bert,bommasani2021opportunities} use tokenization schemes, like BPE or WordPiece tokenization \citep{sennrich-etal-2016-neural,schuster2012japanese,kudo-richardson-2018-sentencepiece}, that break text into word pieces.
These models face an apparent limitation in that they do not have access to information below the level of the word piece, such as information about characters.
But character-level information has been claimed to be useful for a variety of tasks, including adapting text to novel domains like biomedicine, texts with misspellings, and wordplay-based tasks that require attention to character-level manipulations \citep{riabi-etal-2021-character,el-boukkouri-2020-entrainer,clark2021canine}.

There are drawbacks, however, to using character-level models: character-based sequences are long and therefore can slow down training \citep{mielke2021between}.
And including character-level information does not necessarily improve performance on tasks where one might expect it to \citep{libovicky2021don,rosales-nunez-etal-2021-noisy,itzhak2021models}.
Therefore, the vast majority of top-performing models in languages with alphabetic scripts use models with various kinds of subword tokenization schemes \citep[e.g.,][]{devlin-etal-2019-bert,brown2020language}, but rarely with character-level schemes.

One possible explanation for this state of affairs is that models trained on word pieces implicitly learn something about characters, making the explicit inclusion of character-level information unnecessary. 
Indeed, recent work has shown that even models based on subword tokens might be able to use and manipulate character-level information.
\citet{rozner2021} and \citet{efrat-etal-2021-cryptonite} both study cryptic crosswords and find that PLMs (specifically, T5) can  take advantage of character-level information in order to solve wordplay tasks like unscrambling scrambled words.  \citet{itzhak2021models} show that RoBERTa can access subword information by testing it on a spelling task that requires it to map from words to characters (e.g., from \textit{cat} to the characters \textit{c} + \textit{a} + \textit{t}). 

The fact that models can do tasks like this is curious: word pieces have no explicit access to character information during training, and the mechanism by which they acquire such information is not obvious.
The goal of this paper is to understand the nature of this information, and how it is learned.

Thus, we make several contributions.
First, we provide a thorough characterization of what character information is accessible to subword-tokenized PLMs by designing a binary probing task (\S\ref{sec:probing_expt1}) to probe subword tokens for the presence or absence of a particular character: e.g., does the sequence \textit{star} contain the letter \textit{t}? 
This task lets us not just assess whether this information is available, but lets us characterize, in a fine-grained way, the nature of character-level knowledge in subword tokens.
Performance on the task far exceeds a random control as well as a baseline using fixed GloVe word embeddings (an F1 score of 93.7 for the best-performing model, GPT-J), suggesting that subwords learn meaningful information about their characters. 
This result holds across several alphabets (Latin, Devanagari, Cyrillic, Arabic).

To explore how this information is acquired, we introduce several possible explanations and conduct detailed analyses of the probing task on the monolingual English models, with a particular focus on the best-performing model GPT-J (\S\ref{sec:expt1_result_breakdown}). 
Specifically, we consider how character knowledge varies as a function of the character being probed for (it's easier to classify rare letters than common ones), the position in the token of the character in question (performance is somewhat better early in tokens), and the frequency of the token (frequent tokens aren't necessarily easier to probe).
We then turn to the possibility that systematic correspondences between characters and syntactic features (e.g., adverbs tend to end in "y"), play a role in how models acquire character-level information.
To that end, we devise syntactic baselines, whereby we use features like part of speech as input to the classifer for detecting the presence or absence of tokens (\S\ref{sec:expt2_pos_experiment}).
The syntactic probe performs much better than controls, which suggests syntactic features contribute to the tokenizer's performance.
However, this correlation does not suffice to explain the totality of character information learned by PLMs.

Finally, we consider another possible mechanism, based on the variability of tokenization, by which character-level information might be learned (\S\ref{sec:tokenization_variability}). 
We conduct an experiment using simple fixed embeddings, as proof of concept that increasing variability in tokenization \citep{cao-rimell-2021-evaluate} affects the character information learned. 
Overall, given the importance of tokenization schemes for downstream performance \citep{bostrom-etal-2021-flexible, mielke2021between}, we believe richer knowledge as to how tokens acquire character-level information could inform the development of tokenization schemes that improve model performance.

\section{Prior work}

All language models must choose what to use as the basic linguistic unit, and, as a result, there is a long history of work in NLP, evaluating the tradeoffs between models that tokenize words based on characters, words, or something in between, like bytes or word pieces \citep[see][for recent surveys]{mielke2021between,pinter2021integrating}. 

While words are a seemingly natural kind and are often used as basic units for modeling language, there is considerable debate in the linguistics literature as to how to even define  a word, due to differences across languages \citep{martin2017indeterminacy}.
Moreover, word-level models have a major weakness in that they do not naturally handle out of vocabulary items \citep[see][for an overview]{jurafsky2003probabilistic} and can have very different behaviors in languages with different morphological systems \citep{mielke-etal-2019-kind,cotterell2018complexity}.
Character-level models have their own weaknesses: they are typically slower to train at the scale required for massive language modeling.
Many recent efforts have centered around trying to use meaningful sub-word units in language modeling, such as BPE \citep{gage1994new,sennrich-etal-2016-neural}, WordPiece tokenization \citep{schuster2012japanese}, and UnigramLM \citep{kudo-2018-subword}.

While subword tokenization schemes often end up with reasonable linguistic units, they still lack access to character-level information.
So there have been a number of efforts to imbue word or sub-word tokenization schemes with character-level information \citep{mielke2019spell,kim2016character,dos2014learning,bojanowski2017enriching,li2018subword,ma-hovy-2016-end,aguilar-etal-2021-char2subword-extending,el-boukkouri-2020-entrainer,clark2021canine}.

Here, rather than asking how to augment sub-word tokenization schemes with additional information, we ask what they \textit{already} learn about characters naturally.
To do so, we use probing, which is widely used to assess what information is contained in PLM embeddings \citep{belinkov2022probing,belinkov-glass-2019-analysis,hewitt-manning-2019-structural,hupkes2018visualisation}.
Because probing has limitations \citep{elazar-etal-2021-amnesic,pimentel-etal-2020-information,voita-etal-2021-analyzing}, we include a number of control tasks \citep{hewitt-liang-2019-designing} and baselines in order to ask what can be recovered from embeddings, relative to a control of equal expressive power.

\section{Experiment 1: Probing for character information}\label{sec:probing_expt1}

The main goal of our first experiment is to quantify the extent to which tokens in PLMs capture character-level information and characterize that knowledge across a variety of dimensions.
We train a binary classifier probe that takes as input a token's frozen embeddings from PLMs to predict whether a particular character of the alphabet is contained in that token.
That is, if successful, the probe will predict that \textit{cool} contains an "o" but "cat" does not.
We also consider a task in which the probe must say whether one token (e.g., "coo") is a substring of another token (e.g., "cool"). 
We examine the probe's success as a function of the character being probed for, length of the token being probed, position of the character in the token, and frequency of the token.

\subsection{Method}

We consider the static non-contextualized embeddings of the following English PLMs: GPT-J \citep{wang2021gpt}, GPT-2 \citep{radford2019language}, RoBERTa \citep{liu2019roberta}, BERT \citep[cased and uncased;][]{devlin-etal-2019-bert}, as well as GloVe embeddings \citep{pennington2014glove} and Language-only embeddings of the multimodal LXMERT \citep{tan-bansal-2019-lxmert}. To test the generalizability of our results to other languages, we also considered embeddings from Multilingual BART \cite{liu-etal-2020-multilingual-denoising} and used them to test tokens consisting of only English characters, as well as characters from three other alphabetic scripts: Devanagari, Arabic, and Cyrillic.
See Appendix \ref{appendix:Expt1} for model details.

Each language model has its own vocabulary, consisting of tokens. For our English experiments, We consider only the tokens consisting entirely of characters in the standard English alphabet (a-z), along with the special characters that accompany these tokens, such as preceding whitespace (denoted by \.{G} in the RoBERTa and GPT-family) or symbols denoting continuations of preceding word (`\#\#' in BERT family).
Because Multilingual BART consists of characters from different scripts and because its tokens are not explicitly separated by languages, for our Multilingual BART experiments we consider all tokens that consist exclusively of characters from the target alphabet.\footnote{Note that, because Multilingual BART does not explicitly separate tokens based on language, our experiment compares across \textit{scripts}, as opposed to across languages. For instance, the tokens considered for Arabic can include tokens derived from not just the Arabic language, but also other languages that use the Arabic script like Farsi or Malay.}
We define the target alphabet for each script as the alphabetic characters in each script that occur across at least 250 different tokens.

Our main probing task trains classifiers to detect the presence or absence of each of the target characters $\alpha$ in each token $w_i$ from the filtered-vocabulary $V$. Thus, a separate dataset for each character $\alpha$ is constructed over $V$ as $D'_{\alpha} = \{(w_1, y_1), (w_2, y_2), \dots (w_d, y_d)\}$ where the binary label $y_i$ denotes whether $\alpha$ occurs at least once in $w_i \in V$. From these data-points in $D'_{\alpha}$ we create a balanced dataset $D_{\alpha}$ with an equal number of positive and negative labels by undersampling the $(w_i,y_i)$ points with $y_i$ as the negative label 
(i.e., when probing for the presence of the character "z", half the tokens will contain "z" even though most tokens in general do not).
We then split $D_{\alpha}$ into training and test splits in a roughly 80-20 ratio, while (for the English experiments) ensuring that tokens with the same lemma appear in the same split. This is the most challenging split, as it prevents the probe from leveraging wordform similarity across words with the same lemma in both training and test \citep{itzhak2021models}.
Because of technical issues defining lemmas in Multilingual BART, we do not enforce this constraint for the Multilingual BART experiments.

We train our probe on the static non-trainable embeddings $E$ of these PLMs. For a data-point $(w_i, y_i)$, the probe receives as input a token $w_i$ with one-hot encoding $x_i$. The probe predicts logits $\hat{y}_i$ by an MLP: $\hat{y}_i = \sigma(MLP_{\alpha}(E^T x_i))$.
In the control task, we consider randomly-initialized non-trainable embeddings instead of the trained embeddings from the PLMs.

\paragraph{Substring Sub-experiment}

As an additional sub-experiment for assessing the generalizability of the task, for the best-performing English-language model (GPT-J), we consider a related substring classification task.
Specifically, we probe GPT-J's embedding to detect whether a token $u$ is a substring of the token $v$.
That is, can it detect that the token "ome" is a substring of "some"?
For this condition, we set up the experiment as before but, rather than attempt to detect the presence or absence of a character, we seek to classify whether a particular token $u_i$ is a substring of another token $v_i$.
To create positive examples, we consider all substrings of $v_i$ that are in the overall vocabulary $V$. 
For each positive example, we sample a token from $V$ of equal character length as $u_i$ which is \textit{not} a substring of $v_i$ in order to create negative examples. 
This creates a balanced set, from which we sample an 80-20 train-test split, ensuring that the superstring token $v_i$ always occurs in the same split. We train the probe as before, with the input as the concatenated embeddings of the two tokens.

\subsection{Results}

\paragraph{English-Language  Character Probing Results} 

\begin{figure}[t]
    \centering
    \includegraphics[width=.95\columnwidth]{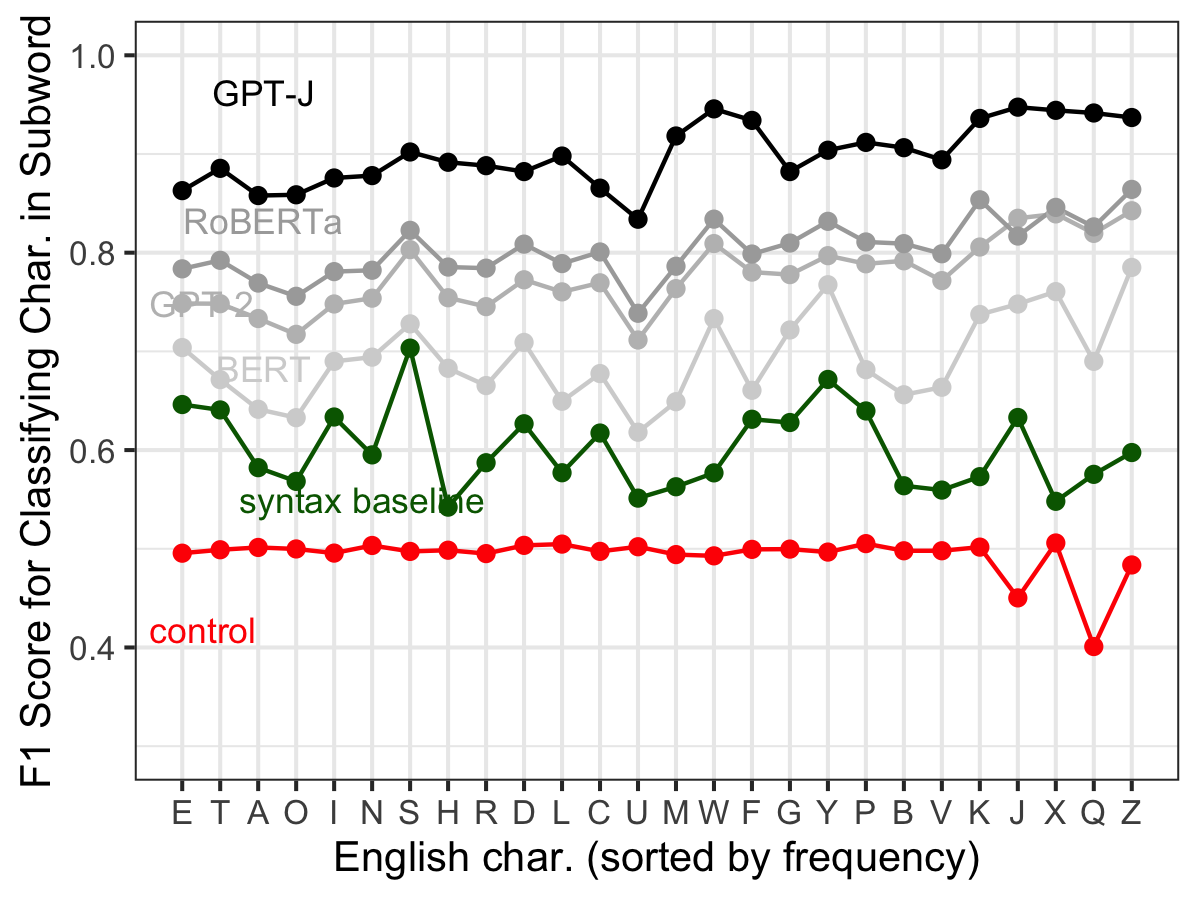}
    \caption{For selected models, the average F1-score (y-axis) for how well a character (x-axis) can be classified on our main probing task. The control (random embeddings) appears in red, the syntax baseline in green. The other 4 models are shown in grayscale, with the largest and most recent model (GPT-J) in the darkest color.}
    \label{fig:page1}
\end{figure}

Table \ref{character_results_mean} shows the results averaged across 5 train-test splits and different seeds, reporting on the Macro-F1 metric averaged across all 26 characters. 
We also observe very low variance for the best-performing models, as shown in the Appendix (Table \ref{character_results_variance}).

For our main character probing experiment, all models perform substantially better than their matched controls (which hover around the chance F! level of 50), suggesting that word piece tokens from PLMs contain information about their constituent characters in their embeddings. 
GPT-J is the best-performing model (with F1 of 93.70 and 94.35), followed by RoBERTa and GPT-2, then the BERT models.
All the transformer models outperform the GloVe fixed embedding model.
Clearly, the performance of the models on this probing task correlates with performance on other language tasks, such that larger models trained on larger corpora do better.\footnote{\noindent Since performance varies considerably based on the model used, we consider this work an additional data point in favor of considering multiple models in interpretability work \citep{bowman2021combating}.}

There are also other factors that may contribute to difference in performance, such as the nature of the pre-training task and the tokenizer. 
The latter is evident from the considerable performance gap between RoBERTa and BERT, which may be partially attributed to RoBERTa using GPT's reversible tokenizer, leading to more variability depending on preceding whitespace. (See  \S\ref{sec:tokenization_variability} for the potential effect of tokenizer variability on performance.)

\begin{table}
 \footnotesize
    \centering
    \begin{tabular}{| c | c | c  |}
    \hline
    Model type & PLM & Control \\
    \hline
    \hline
    \multicolumn{3}{|c|}{\textbf{English Probing Experiment}} \\
    \hline
    GPT-J        & 93.70 & 48.36 \\
    GPT-2        & 84.25 & 52.31 \\
    RoBERTa      & 86.41 & 47.33 \\
    BERT-Cased   & 78.50 & 47.08 \\
    BERT-Uncased & 77.48 & 49.37 \\
    GloVe 300D   & 67.57 & 49.57 \\
    GloVe 100D   & 66.04 & 50.33 \\
    LXMERT       & 62.4 & 53.92 \\
    \hline
    \hline
    \multicolumn{3}{|c|}{\textbf{English Substring Experiment}} \\
    \hline
    GPT-J & 86.56 & 70.03\\
    \hline
    \end{tabular}

    \caption{\label{character_results_mean}
    {Results (F1-scores)for the main English probing experiment.}
    }
\end{table}

\begin{table}
 \footnotesize
    \centering
    \begin{tabular}{| c | c | c  |}
    \hline
     Script & PLM & Control \\
    \hline
     Latin (English chars)      & 80.95 & 39.13 \\
     Devanagari & 78.61 & 50.78 \\
     Arabic     & 76.37 & 51.88 \\
     Cyrillic   & 81.37 & 45.71 \\
    \hline
    \end{tabular}

    \caption{\label{Multilingual_results_mean}
    {Results (F1-scores) for the multilingual probing experiment on Multilingual BART.}
    }
\end{table}

\paragraph{Multilingual Results} Table \ref{Multilingual_results_mean} shows the results for the Multilingual BART experiments, averaged across 5 train-test splits with different seeds. Performance is consistently high and above chance across languages with different scripts. It is highest for Cyrillic with an F1 of 81.37, and lowest for Arabic with an F1 of 76.37.
While we focus mostly on English in the remainder of our experiments because of the large number of different models available and because of the easy access to other sources of linguistic information, we believe these results suggest that our findings would be generalizable to non-Latin scripts. 

\paragraph{English Substring Experiment} Performance on the English Substring Experiment is also far above chance, with an average F1 of 86.56, compared to a control F1 (on random embeddings) of 70.03 (bottom row in Table \ref{character_results_mean}). 
Control performance is well above 50 in this case since the data set is created to be balanced such that the superstrings have equal numbers of positive and negative examples. But there are still baseline differences in how often a token occurs as a substring, so the model can learn that certain substrings like "en" are more common than substrings like "emies". 
We take the performance on the Substring Experiment as evidence that the model can make use of character information to do more complicated substring tasks than just character identification.

\subsection{Breakdown of results} \label{sec:expt1_result_breakdown}

Next, we consider a number of possibilities for how character-level information gets into these embeddings and conduct analyses intended to understand the nature of the information learned and how it gets there.
We focus on our best-performing model (GPT-J) for these analyses.

\paragraph{Is the first letter learned best because of alphabetization?}

\begin{figure}[t]
    \centering
    \includegraphics[width=.94\columnwidth]{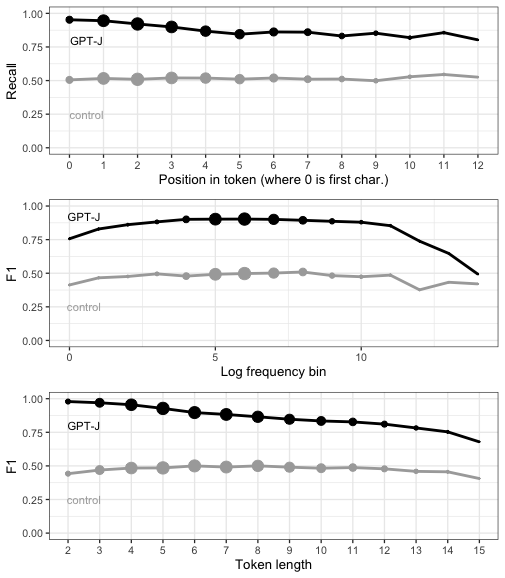}
    \caption{Performance on the GPT-J probe, relative to a control probe, as a function of the character's position in the token (top), the log frequency of the token (middle), and the length of the token (bottom). The size of the point reflects the amount of data.}
    \label{fig:triplot}
\end{figure}

One possibility is that, because the training data likely contains many alphabetical lists and other kinds of word lists (e.g., lists of words starting with "z"), the model learns a co-occurrence relationship between words that start with the same character.
We would predict that this would cause stronger performance when the probed character occurs at the beginning of the word.
To that end, we examine how the model's performance varies as a function of where in the token the target character is (top panel in Figure \ref{fig:triplot}). 
While there is indeed a significant negative relationship between word position and recall as measured by a linear regression ($\beta = -.01$, $p<.001$), the slope is relatively small.
While recall on the first letter in a token is high (95.2), it is not an outlier: performance is only somewhat higher than recall for the second character (94.5).
Moreover, performance is above chance even when the target character appears 10 or more characters deep in a token.
Therefore, we do not believe the effect is driven only by word beginnings, although they likely play a role.

\paragraph{Is it only frequent words that the probe gets right?}

Next, we consider whether performance varies as a function of the frequency of the token (middle panel in Figure \ref{fig:triplot}). 
One possibility could be that character information is memorized only in high-frequency tokens like ``the", which occur often enough that at least sometimes very frequent tokens are broken into characters (e.g., "the" appearing in the context of "t h e"), and that low-frequency tokens will perform worse.
This does not appear to be the case and, in fact, there is, if anything, a negative relationship ($\beta=-.013$, $p=.05$) between binned log frequency and performance, such that less frequent tokens are easier to extract character information from.

\paragraph{Is it easier to get long or short words right?}

The bottom panel of Figure 2 shows F1-score as a function of the length of the token. Using the GPT-J embeddings, it is easier to classify characters in short tokens, as compared to longer tokens. 
This may be a function of the nature of the task since there is, in some sense, less information to be represented for a short token like "be" for the purposes of the task (just that it contains a "b" and it contains an "e"), whereas a long token would have to represent information about more characters. 

\paragraph{Which characters are learned best?}

Part of what makes the success of the probe is that word embeddings represent word co-occurrence information, which is typically conceived of as  semantic in nature \citep{erk2016you} and so should, because of the arbitrariness of the relationship between forms and meanings \citep{saussure1916course,hockett1960origin}, mean there is no relationship between individual characters and information learned by embeddings.
But this arbitrariness breaks down, in that there are statistically detectable non-arbitrary form-meaning relationships in language \citep{blasi2016sound,monaghan_how_2014,tamariz_exploring_2008,dautriche2017words,pimentel-etal-2019-meaning}, such as the fact that \textit{fl-} words in English tend to be about movement \citep[e.g., \textit{flap}, \textit{fly}, \textit{flutter}, \textit{flicker};][]{marchand1959phonetic,bergen2004psychological} and that different parts of speech have different phonological patterns \citep{dautriche2015learning,kelly_using_1992,monaghan_differential_2005}.

An even larger source of shared information between characters and syntactic/semantic information is that morphological forms can be cues to word categories: for instance, most plural nouns end with "s" and many adverbs end in "ly".
This leads to changes in character-level distributions: while roughly 12\% of words in American English contain "y", 85\% of adverbs do \citep[as estimated using data from][]{brysbaert2012adding}.
Thus, a model with access to part of speech information could do well by guessing that all adverbs contain "y".

So one possibility is that the probe's performance is largely driven by characters that correlate with syntactic and semantic features.
If this were the case, we might expect some characters to show much better performance than others.
Figure \ref{fig:page1} shows the F1-Macro as a function of character. For GPT-J, the best-performing model, there are some clear trends.
For instance, it is easiest to classify rare letters: J, W, X, Q, Z all have F1-scores over 93.
And it is hardest for the probe to classify vowels:
U, A, O, and E are the lowest-performing characters, with F1-scores between 83 and 86.
But even those lower-performing characters do far better than the chance baseline (at about 50 F1 score)

To further explore this, we conducted a qualitative analysis of the probe's successes and failures.
Consider the probe for classifying the presence/absence of "y": the model assigns highest confidence to the following 4 tokens: "lly", " selectively", " subtly", " mechanically". These all have "ly" endings, which in English are typically associated with adverbs.
Similarly, the top performing tokens for the "s" classifier all end with a morphologically meaningful "-s" suffix: " socialists", " stocks"," suggestions". 
They also happen to all start with "s", perhaps suggesting an effect of the first character as discussed above.

This analysis suggests that the strong classifier performance could be explained by the model learning systematic relationships between certain characters and syntactically or semantically meaningful morphology. 
Is syntactic information the window through which character-level information enters PLMs? 
To address that question, our next experiment focuses on a syntactic baseline, to see how well character-level information can be predicted based on syntactic features.

\section{Experiment 2: The effect of syntactic information} \label{sec:expt2_pos_experiment}

\begin{table}
 \footnotesize
    \centering
    \begin{tabular}{| c | c | c | c |}
    \hline
    \textbf{Measure} & \textbf{SpaCy} & \textbf{GPT-J} & \textbf{Control} \\
    \hline
    \hline
    \multicolumn{4}{|c|}{\textbf{Aggregate Performance}} \\
    \hline
    F1 & 52.34 & \textbf{61.24} & 49.68 \\
    \hline
    \hline
    \multicolumn{4}{|c|}{\textbf{Best performing characters}} \\
    \hline
    s & 64.60 & \textbf{66.82} & 40.32 \\
    y & 61.96 & \textbf{64.89} & 48.68 \\
    e & 62.05 & \textbf{62.32} & 47.27 \\
    \hline
    \hline
    \multicolumn{4}{|c|}{\textbf{Worst performing characters}} \\
    \hline
    b & 48.92 & \textbf{55.13} & 48.25 \\
    m & 48.13 & \textbf{55.61} & 46.11 \\
    q & 43.79 & \textbf{53.54} & 49.28 \\
    \hline
    \end{tabular}

    \caption{\label{pos_ner_results_main}
    {The best and worst performing characters from Experiment 2 on the SpaCy syntactic baseline, the GPT-J syntactic baseline, and the Control.}
    }
\end{table}

In this experiment, we focus on building probes for the same task as in Experiment 1 (identifying whether a particular character occurs in a particular token). 
But, rather than using the token embeddings from a large language model as input, we attempt to classify the presence/absence of characters in a token based on syntactic information.

Our first model (the SpaCy model) uses the SpaCy library \cite{spacy2} to obtain distributions over features for each token in the vocabulary: Fine-Grained Part of Speech tag (PoS; e.g., for "Jane", NNP for a proper noun), Coarse-Grained Part of Speech tag (Coarse-grained PoS; e.g., for "Jane", PROPN for proper noun), and a Named Entity Recognition tag (NER; e.g., for "Jane", PERSON for a personal name). 
We use these features to construct a syntactic vector for each token.

Because SpaCy is built to operate over words, not tokens, we also construct custom syntactic baselines that can tag subwords, as opposed to tokens.

The performance of these probes will serve as a baseline for ascertaining how much character-level information can be learned by these features alone, without a full language model.
If they can perform just as well as the full GPT-J embeddings, that would suggest that morphosyntactic information (of the sort that we already know is learned by PLMs during pretraining) is sufficient for the performance on the probing task.

The method is the same as in Experiment 1, where the goal is to predict the presence or absence of a character $\alpha$ in a token, except that instead of using the token's model embeddings as input, we instead use syntactic feature vectors (obtained either from SpaCy or a custom tagger) as input. We describe these syntactic vectors below. 

\paragraph{Syntactic baselines}

The SpaCy model has 3 features for each token: NER, PoS, and Coarse-Grained PoS tags. The resultant features are discrete one-hot feature vectors over labels.

The custom syntactic tagger, which is intended to solve the problem that SpaCy tags words and not subword tokens, takes a (subword) token's model embedding as input and outputs a vector of probabilities over part of speech and named entity categories.
Here, we describe results for our custom GPT-J Tagger, trained using GPT-J model embeddings, since GPT-J is the best-performing of our models for our main task.
See Appendix \ref{appendix:Expt2} for descriptions and the results for 2 additional BERT-based custom taggers that we built.

To build our custom GPT-J-Tagger, we train an MLP model to predict PoS and NER labels based on GPT-J's static embedding layer for each token.
The tagger is trained on the CoNLL 2003 dataset's train and evaluation splits \citep{sang2003introduction}, which contain part of speech and named entity information. 
Unlike the SpaCy tagger, our custom GPT-J-Tagger outputs a probability distribution over categories. We use this distribution over labels as input, rather than a one-hot vector. In the Appendix, Table \ref{custom_model_performance} shows the performance of the tagger's performance \textit{qua} tagger.

\paragraph{Probing for characters using syntactic baselines}

We run the character probing experiment as before.
But, rather than using the model embeddings, we use the syntactic feature vectors as the target of our probe.
Table \ref{pos_ner_results_main} shows the results of these experiments.
Using the syntactic baselines leads to substantially improved performance over control tasks, and the GPT-J-Tagger does better than the SpaCy tagger. 
We hypothesize that these divergences occur because the custom GPT-J-Tagger is better suited to handling subwords, and because it enables us to use label distribution rather than one-hot vectors. 

Zooming in on the performance over individual characters, we observe that, relative to the control task, some English characters consistently perform much better when using syntactic features.
As predicted, these are precisely the characters that are highly correlated with particular parts of speech.
The best-performing characters are: "s" (associated with plural nouns and third-person singular verbs) and "y" (associated with adjective and adverb endings).
Thus, the syntactic baselines seem to be capturing the information that they were intended to capture.
But their performance still fell far below the best performing PLMs, suggesting that the large models are capturing more than just the information captured by the syntactic models.
Moreover, as can be seen in Figure \ref{fig:page1}, the syntax baseline shows a sharp peak for morphologically informative characters like "s", but this pattern is much weaker in GPT-J (which shows only a slight performance increase for "s"). 
Therefore, we do not think syntactic information can explain all the character information learned by PLMs.
In the next section, we consider another possibility: variability of tokenization, the focus of the next section.

\section{Experiment 3: Tokenization variability}\label{sec:tokenization_variability}

Consistent with other work suggesting benefits to variable tokenization \citep[e.g.,][]{provilkov-etal-2020-bpe,kudo-2018-subword}, we hypothesize that the variability of tokenization is another avenue by which character-level information could be learned by models.
We first quantify this variability and then run an experiment using CBOW Word Embeddings \citep{mikolov_word2vec_2013a} showing how increasing the variability in tokenization can lead to more character information being learned.
We posit that the same mechanism may be in play for PLMs.

Subword tokenization like the one used by GPT models can cause the same lemma to have very different tokenizations, depending on its form and/or its spelling.
See Table \ref{dictionary} for possible tokenizations of "dictionary" and related forms, including a misspelling (bottom row). 
This is a subset of the possible misspellings, variants, and morphological forms of the word. But the listed forms alone generate 8 unique tokens.

It would be useful for the model to learn a relationship between all these tokens, since they represent the same lemma.
We posit that the desirability of learning this mapping is a mechanism by which character information could be learned, by inducing an objective to map between atomic tokens like "dictionary" and the various substring tokens that can arise.
While each of these mappings could be learned individually, learning character-level spelling information offers a more general solution to the problem, such that even an entirely novel tokenization could be interpreted by composing the characters of the tokens.

\begin{table}
 \footnotesize
    \centering
    \begin{tabular}{| l | l |}
    \hline
    \textbf{Word} & \textbf{Tokenizations} \\
    \hline
    "dictionary" & "d + ictionary" \\
    \hline
    " dictionary" & " dictionary" \\
        \hline
    "dictionaries" & "d + iction + aries"\\
        \hline
    " dictionaries" & " diction + aries" \\
        \hline
    "dicionary" & "d + icion + ary" \\
    \hline
    \end{tabular}
    \caption{Some GPT tokenizations for "dictionary". \label{dictionary}
    }
\end{table}

For this to be plausible, though, variable tokenizations like this must be frequent enough for it to matter.
In Appendix \ref{appendix:Expt3}, we use heuristics to identify different forms in which a word appears and conduct a series of back-of-the-envelope calculations to determine how many different unique tokenizations are expected for a long word (8+ characters) like \textit{dictionary}, in all its variant forms and misspellings in a sample of the Pile corpus  \citep[we used 1/6 of the corpus as a sample;][]{pile_corpus}. We found that, on average, we should expect over 200 different tokenizations for a word like "dictionary", many pairs of which have entirely disjoint sets of subword tokens from each other.

This hypothesis leads to a prediction: increasing the variability of tokenization should increase the amount of character-level information learned. 
To test this, we train models using tokenization schemes with different levels of variability and then test how much character-level information they learn, using our probing task.

\begin{table}
 \footnotesize
    \centering
    \begin{tabular}{| c | c | c | c |}
    \hline
    Tokenization & $\rho$ & Embedding & Control \\
    \hline
    Word  & -    & 60.55 & 47.12 \\
    GPT-J & -    & 63.23 & 47.51 \\
    GPT-J & 0.05 & \textbf{66.00} & 47.23 \\
    GPT-J & 0.1  & 65.64 & 46.72 \\
    GPT-J & 0.2  & 64.23 & 47.01 \\
    GPT-J & 0.5  & 62.33 & 46.47 \\
    \hline
    \end{tabular}

    \caption{Average F1 scores for probing results, as a function of change in tokenization variability \label{variability_expt_results}
    }
\end{table}

Because the overall goal of our paper is to characterize and explain the nature of character-level information learned, we conduct a proof-of-concept experiment using CBOW Word Embeddings \cite{mikolov_word2vec_2013a} on a portion of the Pile corpus with 1.1B characters, as opposed to training a large transformer model from scratch varying tokenization schemes.
We train 6 CBOW models from scratch, each with a different tokenization scheme.
As baselines, we consider vanilla rule-based word-tokenization (the CBOW default, labeled "Word" in Table \ref{variability_expt_results}) and GPT-J's default word piece tokenization scheme.
Comparing these two baselines against each other lets us compare the effect of word tokenization vs. subword tokenization on character information. 
But our key manipulation is to consider variations of GPT-J's tokenizer in which we systematically increase tokenization variability.

In pre-processing the word-tokenized corpus for input, for each word token $w_i$, with probability $(1 - \rho)$, we tokenize it using the standard GPT-J tokenizer. 
Under the standard tokenizer, " schematics" becomes " sche + mat + "ics". 
With probability $\rho$, however, we tokenize $w_i$ using a random tokenization that consists of alternative valid tokens from GPT-J. 
So, " schematics" could become " schema + tics" or " schematic + s" (but not " schemati + cs" since " schemati" is not a valid GPT token).
We vary $\rho$ from 0.05 to 0.5.
See Appendix \ref{appendix:Expt3} for more details on this procedure.
The result is a series of tokenized corpora, which have more variable tokenization than the vanilla GPT-J-tokenized corpus.

We train CBOW models separately for each of these corpora.
Table \ref{variability_expt_results} shows the results of these experiments on our probing task (using the same method as in Experiment 1). 
As expected, probes on the subword tokenization schemes reveal they learn more information about characters than the default word-level tokenizer. 
Most importantly, upon increasing the variability on GPT-J's tokenization scheme, the performance of the probe increases, peaking at $\rho=0.05$ and $\rho=0.1$.
Thereafter, the performance decreases with variability, suggesting that increasing variability leads to increased character knowledge but only up to a point, likely because there is a tradeoff: since the corpus size for the toy experiment is small, having very high variability leads to the model seeing fewer instances of each token.

While the magnitude of these differences is relatively small, they are consistent across random seeds and train-test splits.
Thus, we believe that these results offer proof of concept that the variability of tokenization affects how much character information is learned by CBOW models and that this finding would plausibly generalize to performance in PLMs (although we leave it to future work to confirm this).
As such, increasing tokenization variability could be a means by which PLMs could be engineered to learn richer character-level information.

\section{Discussion and Conclusion}

Overall, our probing methodology revealed that PLMs with sub-word tokenization learn quite a lot about characters.
The result is robust to the position of the character in the token, the identity of the character, the frequency of the token, the length of the token, and the alphabetic script (although we did not consider entirely non-alphabetic scripts like Chinese since such languages would require a very different formulation).

We suggest at least two possible mechanisms by which this information is learned: systematic relationships between certain characters and syntactic/semantic features and the variability of tokenization.
Insofar as these methods (e.g., tokenizer variability) can be manipulated in model construction, this knowledge could be used to build models that perform better at tasks dependent on such knowledge.
Given the particular importance of tokenization in multilingual models \citep{rust-etal-2021-good,singh2019bert}, it would also be fruitful to consider the import of these results for multilingual settings.

More generally, while the linguistic capabilities of PLMs are much studied \citep[for overviews, see][]{rogers-etal-2020-primer,bommasani2021opportunities}, the question whether PLMs learn the constituent characters of tokens is of a different nature in that it depends on learning a property of language (spelling) that is not explicitly tied to meaning. There is no \textit{a priori} reason "dog" is spelled "D-O-G", and, in a sense, the spelling of the word does not matter.
But, in another sense, it \textit{does} matter: humans routinely use language in creative and character-dependent ways: e.g., alphabetizing text, scrambling letters to create codes, and solving crossword puzzles.
Understanding whether and how the building blocks of this meta-linguistic knowledge can emerge during self-supervised training on a word prediction task could be of interest not just in NLP, but in the cognitive sciences.

\section{Ethics and Broader Impacts}

This work consists of probing experiments and interpretability analyses of PLMs, and the risks and ethical considerations are largely those that affect any work with large PLMs \citep[e.g., energy costs; see][for an overview of risks and tradeoffs]{bommasani2021opportunities}. 
The intended use of our code is for academic research.
We consider probing publicly available PLMs, which are made publicly available in part for research purposes, to be within the intended use of PLMs.

\section{Acknowledgments}

This work was supported by National Science Foundation Grants No. 2104995 to KM. 
We thank Chris Potts and Josh Rozner for conversations that helped inspire this work, Maria Ryskina and Kaj Bostrom for comments on drafts, and Eunsol Choi for introducing the authors.

\bibliography{bibliography,anthology2,custom,everything}

\begin{thebibliography}{72}
\expandafter\ifx\csname natexlab\endcsname\relax\def\natexlab#1{#1}\fi

\bibitem[{Aguilar et~al.(2021)Aguilar, McCann, Niu, Rajani, Keskar, and
  Solorio}]{aguilar-etal-2021-char2subword-extending}
Gustavo Aguilar, Bryan McCann, Tong Niu, Nazneen Rajani, Nitish~Shirish Keskar,
  and Thamar Solorio. 2021.
\newblock \href {https://doi.org/10.18653/v1/2021.findings-emnlp.141}
  {{C}har2{S}ubword: Extending the subword embedding space using robust
  character compositionality}.
\newblock In \emph{Findings of the Association for Computational Linguistics:
  EMNLP 2021}, pages 1640--1651, Punta Cana, Dominican Republic. Association
  for Computational Linguistics.

\bibitem[{Belinkov(2022)}]{belinkov2022probing}
Yonatan Belinkov. 2022.
\newblock Probing classifiers: Promises, shortcomings, and advances.
\newblock \emph{Computational Linguistics}, 48(1):207--219.

\bibitem[{Belinkov and Glass(2019)}]{belinkov-glass-2019-analysis}
Yonatan Belinkov and James Glass. 2019.
\newblock \href {https://doi.org/10.1162/tacl_a_00254} {Analysis methods in
  neural language processing: A survey}.
\newblock \emph{Transactions of the Association for Computational Linguistics},
  7:49--72.

\bibitem[{Bergen(2004)}]{bergen2004psychological}
Benjamin~K Bergen. 2004.
\newblock The psychological reality of phonaesthemes.
\newblock \emph{Language}, 80(2):290--311.

\bibitem[{Bird et~al.(2009)Bird, Klein, and Loper}]{bird2009natural}
Steven Bird, Ewan Klein, and Edward Loper. 2009.
\newblock \emph{Natural language processing with Python: analyzing text with
  the natural language toolkit}.
\newblock " O'Reilly Media, Inc.".

\bibitem[{Blasi et~al.(2016)Blasi, Wichmann, Hammarstr{\"o}m, Stadler, and
  Christiansen}]{blasi2016sound}
Dami{\'a}n~E Blasi, S{\o}ren Wichmann, Harald Hammarstr{\"o}m, Peter~F Stadler,
  and Morten~H Christiansen. 2016.
\newblock Sound--meaning association biases evidenced across thousands of
  languages.
\newblock \emph{Proceedings of the National Academy of Sciences},
  113(39):10818--10823.

\bibitem[{Bojanowski et~al.(2017)Bojanowski, Grave, Joulin, and
  Mikolov}]{bojanowski2017enriching}
Piotr Bojanowski, Edouard Grave, Armand Joulin, and Tomas Mikolov. 2017.
\newblock Enriching word vectors with subword information.
\newblock \emph{Transactions of the Association for Computational Linguistics},
  5:135--146.

\bibitem[{Bommasani et~al.(2021)Bommasani, Hudson, Adeli, Altman, Arora, von
  Arx, Bernstein, Bohg, Bosselut, Brunskill
  et~al.}]{bommasani2021opportunities}
Rishi Bommasani, Drew~A Hudson, Ehsan Adeli, Russ Altman, Simran Arora, Sydney
  von Arx, Michael~S Bernstein, Jeannette Bohg, Antoine Bosselut, Emma
  Brunskill, et~al. 2021.
\newblock On the opportunities and risks of foundation models.
\newblock \emph{arXiv preprint arXiv:2108.07258}.

\bibitem[{Bostrom et~al.(2021)Bostrom, Zhao, Chaudhuri, and
  Durrett}]{bostrom-etal-2021-flexible}
Kaj Bostrom, Xinyu Zhao, Swarat Chaudhuri, and Greg Durrett. 2021.
\newblock \href {https://doi.org/10.18653/v1/2021.emnlp-main.506} {Flexible
  generation of natural language deductions}.
\newblock In \emph{Proceedings of the 2021 Conference on Empirical Methods in
  Natural Language Processing}, pages 6266--6278, Online and Punta Cana,
  Dominican Republic. Association for Computational Linguistics.

\bibitem[{Bowman(2021)}]{bowman2021combating}
Samuel~R Bowman. 2021.
\newblock When combating hype, proceed with caution.
\newblock \emph{arXiv preprint arXiv:2110.08300}.

\bibitem[{Brown et~al.(2020)Brown, Mann, Ryder, Subbiah, Kaplan, Dhariwal,
  Neelakantan, Shyam, Sastry, Askell et~al.}]{brown2020language}
Tom~B Brown, Benjamin Mann, Nick Ryder, Melanie Subbiah, Jared Kaplan, Prafulla
  Dhariwal, Arvind Neelakantan, Pranav Shyam, Girish Sastry, Amanda Askell,
  et~al. 2020.
\newblock Language models are few-shot learners.
\newblock \emph{arXiv preprint arXiv:2005.14165}.

\bibitem[{Brysbaert et~al.(2012)Brysbaert, New, and
  Keuleers}]{brysbaert2012adding}
Marc Brysbaert, Boris New, and Emmanuel Keuleers. 2012.
\newblock Adding part-of-speech information to the subtlex-us word frequencies.
\newblock \emph{Behavior research methods}, 44(4):991--997.

\bibitem[{Cao and Rimell(2021)}]{cao-rimell-2021-evaluate}
Kris Cao and Laura Rimell. 2021.
\newblock \href {https://doi.org/10.18653/v1/2021.emnlp-main.161} {You should
  evaluate your language model on marginal likelihood over tokenisations}.
\newblock In \emph{Proceedings of the 2021 Conference on Empirical Methods in
  Natural Language Processing}, pages 2104--2114, Online and Punta Cana,
  Dominican Republic. Association for Computational Linguistics.

\bibitem[{Clark et~al.(2021)Clark, Garrette, Turc, and
  Wieting}]{clark2021canine}
Jonathan~H Clark, Dan Garrette, Iulia Turc, and John Wieting. 2021.
\newblock Canine: Pre-training an efficient tokenization-free encoder for
  language representation.
\newblock \emph{arXiv preprint arXiv:2103.06874}.

\bibitem[{Cotterell et~al.(2018)Cotterell, Kirov, Hulden, and
  Eisner}]{cotterell2018complexity}
Ryan Cotterell, Christo Kirov, Mans Hulden, and Jason Eisner. 2018.
\newblock On the complexity and typology of inflectional morphological systems.
\newblock \emph{Transactions of the Association for Computational Linguistics}.

\bibitem[{Dautriche et~al.(2017)Dautriche, Mahowald, Gibson, Christophe, and
  Piantadosi}]{dautriche2017words}
Isabelle Dautriche, Kyle Mahowald, Edward Gibson, Anne Christophe, and Steven~T
  Piantadosi. 2017.
\newblock Words cluster phonetically beyond phonotactic regularities.
\newblock \emph{Cognition}, 163:128--145.

\bibitem[{Dautriche et~al.(2015)Dautriche, Swingley, and
  Christophe}]{dautriche2015learning}
Isabelle Dautriche, Daniel Swingley, and Anne Christophe. 2015.
\newblock Learning novel phonological neighbors: Syntactic category matters.
\newblock \emph{Cognition}, 143:77--86.

\bibitem[{Devlin et~al.(2019)Devlin, Chang, Lee, and
  Toutanova}]{devlin-etal-2019-bert}
Jacob Devlin, Ming-Wei Chang, Kenton Lee, and Kristina Toutanova. 2019.
\newblock \href {https://doi.org/10.18653/v1/N19-1423} {{BERT}: Pre-training of
  deep bidirectional transformers for language understanding}.
\newblock In \emph{Proceedings of the 2019 Conference of the North {A}merican
  Chapter of the Association for Computational Linguistics: Human Language
  Technologies, Volume 1 (Long and Short Papers)}, pages 4171--4186,
  Minneapolis, Minnesota. Association for Computational Linguistics.

\bibitem[{Dos~Santos and Zadrozny(2014)}]{dos2014learning}
Cicero Dos~Santos and Bianca Zadrozny. 2014.
\newblock Learning character-level representations for part-of-speech tagging.
\newblock In \emph{International Conference on Machine Learning}, pages
  1818--1826. PMLR.

\bibitem[{Efrat et~al.(2021)Efrat, Shaham, Kilman, and
  Levy}]{efrat-etal-2021-cryptonite}
Avia Efrat, Uri Shaham, Dan Kilman, and Omer Levy. 2021.
\newblock \href {https://doi.org/10.18653/v1/2021.emnlp-main.344} {Cryptonite:
  A cryptic crossword benchmark for extreme ambiguity in language}.
\newblock In \emph{Proceedings of the 2021 Conference on Empirical Methods in
  Natural Language Processing}, pages 4186--4192, Online and Punta Cana,
  Dominican Republic. Association for Computational Linguistics.

\bibitem[{El~Boukkouri(2020)}]{el-boukkouri-2020-entrainer}
Hicham El~Boukkouri. 2020.
\newblock \href {https://aclanthology.org/2020.jeptalnrecital-recital.3}
  {R{\'e}-entra{\^\i}ner ou entra{\^\i}ner soi-m{\^e}me ? strat{\'e}gies de
  pr{\'e}-entra{\^\i}nement de {BERT} en domaine m{\'e}dical (re-train or train
  from scratch ? pre-training strategies for {BERT} in the medical domain )}.
\newblock In \emph{Actes de la 6e conf{\'e}rence conjointe Journ{\'e}es
  d'{\'E}tudes sur la Parole (JEP, 33e {\'e}dition), Traitement Automatique des
  Langues Naturelles (TALN, 27e {\'e}dition), Rencontre des {\'E}tudiants
  Chercheurs en Informatique pour le Traitement Automatique des Langues
  (R{\'E}CITAL, 22e {\'e}dition). Volume 3 : Rencontre des {\'E}tudiants
  Chercheurs en Informatique pour le TAL}, pages 29--42, Nancy, France. ATALA
  et AFCP.

\bibitem[{Elazar et~al.(2021)Elazar, Ravfogel, Jacovi, and
  Goldberg}]{elazar-etal-2021-amnesic}
Yanai Elazar, Shauli Ravfogel, Alon Jacovi, and Yoav Goldberg. 2021.
\newblock \href {https://doi.org/10.1162/tacl_a_00359} {Amnesic probing:
  Behavioral explanation with amnesic counterfactuals}.
\newblock \emph{Transactions of the Association for Computational Linguistics},
  9:160--175.

\bibitem[{Erk(2016)}]{erk2016you}
Katrin Erk. 2016.
\newblock What do you know about an alligator when you know the company it
  keeps?
\newblock \emph{Semantics and Pragmatics}, 9:17--1.

\bibitem[{Gage(1994)}]{gage1994new}
Philip Gage. 1994.
\newblock A new algorithm for data compression.
\newblock \emph{C Users Journal}, 12(2):23--38.

\bibitem[{Gao et~al.(2020)Gao, Biderman, Black, Golding, Hoppe, Foster, Phang,
  He, Thite, Nabeshima, Presser, and Leahy}]{pile_corpus}
Leo Gao, Stella Biderman, Sid Black, Laurence Golding, Travis Hoppe, Charles
  Foster, Jason Phang, Horace He, Anish Thite, Noa Nabeshima, Shawn Presser,
  and Connor Leahy. 2020.
\newblock \href {http://arxiv.org/abs/arXiv:2101.00027} {The pile: An 800gb
  dataset of diverse text for language modeling}.

\bibitem[{Haspelmath(2017)}]{martin2017indeterminacy}
Martin Haspelmath. 2017.
\newblock The indeterminacy of word segmentation and the nature of morphology
  and syntax.
\newblock \emph{Folia {L}inguistica}, 51(s1000):31--80.

\bibitem[{Hewitt and Liang(2019)}]{hewitt-liang-2019-designing}
John Hewitt and Percy Liang. 2019.
\newblock \href {https://doi.org/10.18653/v1/D19-1275} {Designing and
  interpreting probes with control tasks}.
\newblock In \emph{Proceedings of the 2019 Conference on Empirical Methods in
  Natural Language Processing and the 9th International Joint Conference on
  Natural Language Processing (EMNLP-IJCNLP)}, pages 2733--2743, Hong Kong,
  China. Association for Computational Linguistics.

\bibitem[{Hewitt and Manning(2019)}]{hewitt-manning-2019-structural}
John Hewitt and Christopher~D. Manning. 2019.
\newblock \href {https://doi.org/10.18653/v1/N19-1419} {{A} structural probe
  for finding syntax in word representations}.
\newblock In \emph{Proceedings of the 2019 Conference of the North {A}merican
  Chapter of the Association for Computational Linguistics: Human Language
  Technologies, Volume 1 (Long and Short Papers)}, pages 4129--4138,
  Minneapolis, Minnesota. Association for Computational Linguistics.

\bibitem[{Hockett(1960)}]{hockett1960origin}
C.F. Hockett. 1960.
\newblock The origin of language.
\newblock \emph{Scientific American}, 203(3):88--96.

\bibitem[{Honnibal and Montani(2017)}]{spacy2}
Matthew Honnibal and Ines Montani. 2017.
\newblock {spaCy 2}: Natural language understanding with {B}loom embeddings,
  convolutional neural networks and incremental parsing.
\newblock To appear.

\bibitem[{Hupkes et~al.(2018)Hupkes, Veldhoen, and
  Zuidema}]{hupkes2018visualisation}
Dieuwke Hupkes, Sara Veldhoen, and Willem Zuidema. 2018.
\newblock Visualisation and 'diagnostic classifiers' reveal how recurrent and
  recursive neural networks process hierarchical structure.
\newblock \emph{Journal of Artificial Intelligence Research}, 61:907--926.

\bibitem[{Itzhak and Levy(2021)}]{itzhak2021models}
Itay Itzhak and Omer Levy. 2021.
\newblock Models in a spelling bee: Language models implicitly learn the
  character composition of tokens.
\newblock \emph{arXiv preprint arXiv:2108.11193}.

\bibitem[{Jurafsky(2003)}]{jurafsky2003probabilistic}
Daniel Jurafsky. 2003.
\newblock Probabilistic modeling in psycholinguistics: {L}inguistic
  comprehension and production.
\newblock In R.~Bod, J.~Hay, and S.~Jannedy, editors, \emph{Probabilistic
  Linguistics}. MIT Press.

\bibitem[{Kelly(1992)}]{kelly_using_1992}
Michael~H. Kelly. 1992.
\newblock \href {https://doi.org/10.1037/0033-295X.99.2.349} {Using sound to
  solve syntactic problems: The role of phonology in grammatical category
  assignments.}
\newblock \emph{Psychological Review}, 99(2):349--364.

\bibitem[{Kim et~al.(2016)Kim, Jernite, Sontag, and Rush}]{kim2016character}
Yoon Kim, Yacine Jernite, David Sontag, and Alexander~M Rush. 2016.
\newblock Character-aware neural language models.
\newblock In \emph{Thirtieth AAAI conference on artificial intelligence}.

\bibitem[{Kingma and Ba(2015)}]{kingma}
Diederik~P. Kingma and Jimmy Ba. 2015.
\newblock \href {http://arxiv.org/abs/1412.6980} {Adam: {A} method for
  stochastic optimization}.
\newblock In \emph{3rd International Conference on Learning Representations,
  {ICLR} 2015, San Diego, CA, USA, May 7-9, 2015, Conference Track
  Proceedings}.

\bibitem[{Kudo(2018)}]{kudo-2018-subword}
Taku Kudo. 2018.
\newblock \href {https://doi.org/10.18653/v1/P18-1007} {Subword regularization:
  Improving neural network translation models with multiple subword
  candidates}.
\newblock In \emph{Proceedings of the 56th Annual Meeting of the Association
  for Computational Linguistics (Volume 1: Long Papers)}, pages 66--75,
  Melbourne, Australia. Association for Computational Linguistics.

\bibitem[{Kudo and Richardson(2018)}]{kudo-richardson-2018-sentencepiece}
Taku Kudo and John Richardson. 2018.
\newblock \href {https://doi.org/10.18653/v1/D18-2012} {{S}entence{P}iece: A
  simple and language independent subword tokenizer and detokenizer for neural
  text processing}.
\newblock In \emph{Proceedings of the 2018 Conference on Empirical Methods in
  Natural Language Processing: System Demonstrations}, pages 66--71, Brussels,
  Belgium. Association for Computational Linguistics.

\bibitem[{Li et~al.(2018)Li, Drozd, Liu, and Du}]{li2018subword}
Bofang Li, Aleksandr Drozd, Tao Liu, and Xiaoyong Du. 2018.
\newblock Subword-level composition functions for learning word embeddings.
\newblock In \emph{Proceedings of the second workshop on subword/character
  level models}, pages 38--48.

\bibitem[{Libovick{\`y} et~al.(2021)Libovick{\`y}, Schmid, and
  Fraser}]{libovicky2021don}
Jind{\v{r}}ich Libovick{\`y}, Helmut Schmid, and Alexander Fraser. 2021.
\newblock Why don't people use character-level machine translation?
\newblock \emph{arXiv preprint arXiv:2110.08191}.

\bibitem[{Liu et~al.(2020)Liu, Gu, Goyal, Li, Edunov, Ghazvininejad, Lewis, and
  Zettlemoyer}]{liu-etal-2020-multilingual-denoising}
Yinhan Liu, Jiatao Gu, Naman Goyal, Xian Li, Sergey Edunov, Marjan
  Ghazvininejad, Mike Lewis, and Luke Zettlemoyer. 2020.
\newblock \href {https://doi.org/10.1162/tacl_a_00343} {Multilingual denoising
  pre-training for neural machine translation}.
\newblock \emph{Transactions of the Association for Computational Linguistics},
  8:726--742.

\bibitem[{Liu et~al.(2019)Liu, Ott, Goyal, Du, Joshi, Chen, Levy, Lewis,
  Zettlemoyer, and Stoyanov}]{liu2019roberta}
Yinhan Liu, Myle Ott, Naman Goyal, Jingfei Du, Mandar Joshi, Danqi Chen, Omer
  Levy, Mike Lewis, Luke Zettlemoyer, and Veselin Stoyanov. 2019.
\newblock Roberta: A robustly optimized bert pretraining approach.
\newblock \emph{arXiv preprint arXiv:1907.11692}.

\bibitem[{Ma and Hovy(2016)}]{ma-hovy-2016-end}
Xuezhe Ma and Eduard Hovy. 2016.
\newblock \href {https://doi.org/10.18653/v1/P16-1101} {End-to-end sequence
  labeling via bi-directional {LSTM}-{CNN}s-{CRF}}.
\newblock In \emph{Proceedings of the 54th Annual Meeting of the Association
  for Computational Linguistics (Volume 1: Long Papers)}, pages 1064--1074,
  Berlin, Germany. Association for Computational Linguistics.

\bibitem[{Marchand(1959)}]{marchand1959phonetic}
Hans Marchand. 1959.
\newblock Phonetic symbolism in english wordformation.
\newblock \emph{Indogermanische Forschungen}, 64:146.

\bibitem[{Mielke et~al.(2021)Mielke, Alyafeai, Salesky, Raffel, Dey, Gall{\'e},
  Raja, Si, Lee, Sagot et~al.}]{mielke2021between}
Sabrina~J Mielke, Zaid Alyafeai, Elizabeth Salesky, Colin Raffel, Manan Dey,
  Matthias Gall{\'e}, Arun Raja, Chenglei Si, Wilson~Y Lee, Beno{\^\i}t Sagot,
  et~al. 2021.
\newblock Between words and characters: A brief history of open-vocabulary
  modeling and tokenization in nlp.
\newblock \emph{arXiv preprint arXiv:2112.10508}.

\bibitem[{Mielke et~al.(2019)Mielke, Cotterell, Gorman, Roark, and
  Eisner}]{mielke-etal-2019-kind}
Sabrina~J. Mielke, Ryan Cotterell, Kyle Gorman, Brian Roark, and Jason Eisner.
  2019.
\newblock \href {https://doi.org/10.18653/v1/P19-1491} {What kind of language
  is hard to language-model?}
\newblock In \emph{Proceedings of the 57th Annual Meeting of the Association
  for Computational Linguistics}, pages 4975--4989, Florence, Italy.
  Association for Computational Linguistics.

\bibitem[{Mielke and Eisner(2019)}]{mielke2019spell}
Sabrina~J Mielke and Jason Eisner. 2019.
\newblock Spell once, summon anywhere: A two-level open-vocabulary language
  model.
\newblock In \emph{Proceedings of the AAAI Conference on Artificial
  Intelligence}, volume~33, pages 6843--6850.

\bibitem[{Mikolov et~al.(2013)Mikolov, Chen, Corrado, and
  Dean}]{mikolov_word2vec_2013a}
Tomas Mikolov, Kai Chen, Greg Corrado, and Jeffrey Dean. 2013.
\newblock \href {http://arxiv.org/abs/arXiv:1301.3781} {Efficient estimation of
  word representations in vector space}.

\bibitem[{Monaghan et~al.(2005)Monaghan, Chater, and
  Christiansen}]{monaghan_differential_2005}
Padraic Monaghan, Nick Chater, and Morten~H. Christiansen. 2005.
\newblock \href {https://doi.org/10.1016/j.cognition.2004.09.001} {The
  differential role of phonological and distributional cues in grammatical
  categorisation}.
\newblock \emph{Cognition}, 96(2):143--182.

\bibitem[{Monaghan et~al.(2014)Monaghan, Shillcock, Christiansen, and
  Kirby}]{monaghan_how_2014}
Padraic Monaghan, Richard~C. Shillcock, Morten~H. Christiansen, and Simon
  Kirby. 2014.
\newblock How arbitrary is language.
\newblock \emph{Philosophical Transactions of the Royal Society B}.

\bibitem[{Paszke et~al.(2019)Paszke, Gross, Massa, Lerer, Bradbury, Chanan,
  Killeen, Lin, Gimelshein, Antiga, Desmaison, K{\"{o}}pf, Yang, DeVito,
  Raison, Tejani, Chilamkurthy, Steiner, Fang, Bai, and
  Chintala}]{pytorch_paper}
Adam Paszke, Sam Gross, Francisco Massa, Adam Lerer, James Bradbury, Gregory
  Chanan, Trevor Killeen, Zeming Lin, Natalia Gimelshein, Luca Antiga, Alban
  Desmaison, Andreas K{\"{o}}pf, Edward~Z. Yang, Zachary DeVito, Martin Raison,
  Alykhan Tejani, Sasank Chilamkurthy, Benoit Steiner, Lu~Fang, Junjie Bai, and
  Soumith Chintala. 2019.
\newblock \href
  {https://proceedings.neurips.cc/paper/2019/hash/bdbca288fee7f92f2bfa9f7012727740-Abstract.html}
  {Pytorch: An imperative style, high-performance deep learning library}.
\newblock In \emph{Advances in Neural Information Processing Systems 32: Annual
  Conference on Neural Information Processing Systems 2019, NeurIPS 2019,
  December 8-14, 2019, Vancouver, BC, Canada}, pages 8024--8035.

\bibitem[{Pennington et~al.(2014)Pennington, Socher, and
  Manning}]{pennington2014glove}
Jeffrey Pennington, Richard Socher, and Christopher~D Manning. 2014.
\newblock Glove: Global vectors for word representation.
\newblock In \emph{Proceedings of the 2014 conference on empirical methods in
  natural language processing (EMNLP)}, pages 1532--1543.

\bibitem[{Pimentel et~al.(2019)Pimentel, McCarthy, Blasi, Roark, and
  Cotterell}]{pimentel-etal-2019-meaning}
Tiago Pimentel, Arya~D. McCarthy, Damian Blasi, Brian Roark, and Ryan
  Cotterell. 2019.
\newblock \href {https://doi.org/10.18653/v1/P19-1171} {Meaning to form:
  Measuring systematicity as information}.
\newblock In \emph{Proceedings of the 57th Annual Meeting of the Association
  for Computational Linguistics}, pages 1751--1764, Florence, Italy.
  Association for Computational Linguistics.

\bibitem[{Pimentel et~al.(2020)Pimentel, Valvoda, Hall~Maudslay, Zmigrod,
  Williams, and Cotterell}]{pimentel-etal-2020-information}
Tiago Pimentel, Josef Valvoda, Rowan Hall~Maudslay, Ran Zmigrod, Adina
  Williams, and Ryan Cotterell. 2020.
\newblock \href {https://doi.org/10.18653/v1/2020.acl-main.420}
  {Information-theoretic probing for linguistic structure}.
\newblock In \emph{Proceedings of the 58th Annual Meeting of the Association
  for Computational Linguistics}, pages 4609--4622, Online. Association for
  Computational Linguistics.

\bibitem[{Pinter(2021)}]{pinter2021integrating}
Yuval Pinter. 2021.
\newblock Integrating approaches to word representation.
\newblock \emph{arXiv preprint arXiv:2109.04876}.

\bibitem[{Provilkov et~al.(2020)Provilkov, Emelianenko, and
  Voita}]{provilkov-etal-2020-bpe}
Ivan Provilkov, Dmitrii Emelianenko, and Elena Voita. 2020.
\newblock \href {https://doi.org/10.18653/v1/2020.acl-main.170} {{BPE}-dropout:
  Simple and effective subword regularization}.
\newblock In \emph{Proceedings of the 58th Annual Meeting of the Association
  for Computational Linguistics}, pages 1882--1892, Online. Association for
  Computational Linguistics.

\bibitem[{Radford et~al.(2019)Radford, Wu, Child, Luan, Amodei, Sutskever
  et~al.}]{radford2019language}
Alec Radford, Jeffrey Wu, Rewon Child, David Luan, Dario Amodei, Ilya
  Sutskever, et~al. 2019.
\newblock Language models are unsupervised multitask learners.
\newblock \emph{OpenAI blog}, 1(8):9.

\bibitem[{Riabi et~al.(2021)Riabi, Sagot, and
  Seddah}]{riabi-etal-2021-character}
Arij Riabi, Beno{\^\i}t Sagot, and Djam{\'e} Seddah. 2021.
\newblock \href {https://doi.org/10.18653/v1/2021.wnut-1.47} {Can
  character-based language models improve downstream task performances in
  low-resource and noisy language scenarios?}
\newblock In \emph{Proceedings of the Seventh Workshop on Noisy User-generated
  Text (W-NUT 2021)}, pages 423--436, Online. Association for Computational
  Linguistics.

\bibitem[{Rogers et~al.(2020)Rogers, Kovaleva, and
  Rumshisky}]{rogers-etal-2020-primer}
Anna Rogers, Olga Kovaleva, and Anna Rumshisky. 2020.
\newblock \href {https://doi.org/10.1162/tacl_a_00349} {A primer in
  {BERT}ology: What we know about how {BERT} works}.
\newblock \emph{Transactions of the Association for Computational Linguistics},
  8:842--866.

\bibitem[{Rosales~N{\'u}{\~n}ez et~al.(2021)Rosales~N{\'u}{\~n}ez, Wisniewski,
  and Seddah}]{rosales-nunez-etal-2021-noisy}
Jos{\'e}~Carlos Rosales~N{\'u}{\~n}ez, Guillaume Wisniewski, and Djam{\'e}
  Seddah. 2021.
\newblock \href {https://doi.org/10.18653/v1/2021.wnut-1.23} {Noisy {UGC}
  translation at the character level: Revisiting open-vocabulary capabilities
  and robustness of char-based models}.
\newblock In \emph{Proceedings of the Seventh Workshop on Noisy User-generated
  Text (W-NUT 2021)}, pages 199--211, Online. Association for Computational
  Linguistics.

\bibitem[{Rozner et~al.(2021)Rozner, Potts, and Mahowald}]{rozner2021}
Josh Rozner, Christopher Potts, and Kyle Mahowald. 2021.
\newblock \href
  {https://proceedings.neurips.cc/paper/2021/file/5f1d3986fae10ed2994d14ecd89892d7-Paper.pdf}
  {Decrypting cryptic crosswords: Semantically complex wordplay puzzles as a
  target for nlp}.
\newblock In \emph{Advances in Neural Information Processing Systems},
  volume~34, pages 11409--11421. Curran Associates, Inc.

\bibitem[{Rust et~al.(2021)Rust, Pfeiffer, Vuli{\'c}, Ruder, and
  Gurevych}]{rust-etal-2021-good}
Phillip Rust, Jonas Pfeiffer, Ivan Vuli{\'c}, Sebastian Ruder, and Iryna
  Gurevych. 2021.
\newblock \href {https://doi.org/10.18653/v1/2021.acl-long.243} {How good is
  your tokenizer? on the monolingual performance of multilingual language
  models}.
\newblock In \emph{Proceedings of the 59th Annual Meeting of the Association
  for Computational Linguistics and the 11th International Joint Conference on
  Natural Language Processing (Volume 1: Long Papers)}, pages 3118--3135,
  Online. Association for Computational Linguistics.

\bibitem[{Sang and De~Meulder(2003)}]{sang2003introduction}
Erik~F Sang and Fien De~Meulder. 2003.
\newblock Introduction to the conll-2003 shared task: Language-independent
  named entity recognition.
\newblock \emph{arXiv preprint cs/0306050}.

\bibitem[{Saussure(1916)}]{saussure1916course}
F.~de Saussure. 1916.
\newblock \emph{Course in general linguistics}.
\newblock Open Court Publishing Company.

\bibitem[{Schuster and Nakajima(2012)}]{schuster2012japanese}
Mike Schuster and Kaisuke Nakajima. 2012.
\newblock Japanese and korean voice search.
\newblock In \emph{2012 IEEE International Conference on Acoustics, Speech and
  Signal Processing (ICASSP)}, pages 5149--5152. IEEE.

\bibitem[{Sennrich et~al.(2016)Sennrich, Haddow, and
  Birch}]{sennrich-etal-2016-neural}
Rico Sennrich, Barry Haddow, and Alexandra Birch. 2016.
\newblock \href {https://doi.org/10.18653/v1/P16-1162} {Neural machine
  translation of rare words with subword units}.
\newblock In \emph{Proceedings of the 54th Annual Meeting of the Association
  for Computational Linguistics (Volume 1: Long Papers)}, pages 1715--1725,
  Berlin, Germany. Association for Computational Linguistics.

\bibitem[{Singh et~al.(2019)Singh, McCann, Socher, and Xiong}]{singh2019bert}
Jasdeep Singh, Bryan McCann, Richard Socher, and Caiming Xiong. 2019.
\newblock Bert is not an interlingua and the bias of tokenization.
\newblock In \emph{Proceedings of the 2nd Workshop on Deep Learning Approaches
  for Low-Resource NLP (DeepLo 2019)}, pages 47--55.

\bibitem[{Tamariz(2008)}]{tamariz_exploring_2008}
Monica Tamariz. 2008.
\newblock \href {https://doi.org/10.1075/ml.3.2.05tam} {Exploring systematicity
  between phonological and context-cooccurrence representations of the mental
  lexicon}.
\newblock \emph{The Mental Lexicon}, 3(2):259--278.

\bibitem[{Tan and Bansal(2019)}]{tan-bansal-2019-lxmert}
Hao Tan and Mohit Bansal. 2019.
\newblock \href {https://doi.org/10.18653/v1/D19-1514} {{LXMERT}: Learning
  cross-modality encoder representations from transformers}.
\newblock In \emph{Proceedings of the 2019 Conference on Empirical Methods in
  Natural Language Processing and the 9th International Joint Conference on
  Natural Language Processing (EMNLP-IJCNLP)}, pages 5100--5111, Hong Kong,
  China. Association for Computational Linguistics.

\bibitem[{Voita et~al.(2021)Voita, Sennrich, and
  Titov}]{voita-etal-2021-analyzing}
Elena Voita, Rico Sennrich, and Ivan Titov. 2021.
\newblock \href {https://doi.org/10.18653/v1/2021.acl-long.91} {Analyzing the
  source and target contributions to predictions in neural machine
  translation}.
\newblock In \emph{Proceedings of the 59th Annual Meeting of the Association
  for Computational Linguistics and the 11th International Joint Conference on
  Natural Language Processing (Volume 1: Long Papers)}, pages 1126--1140,
  Online. Association for Computational Linguistics.

\bibitem[{Wang and Komatsuzaki(2021)}]{wang2021gpt}
Ben Wang and Aran Komatsuzaki. 2021.
\newblock Gpt-j-6b: A 6 billion parameter autoregressive language model.

\bibitem[{Wolf et~al.(2019)Wolf, Debut, Sanh, Chaumond, Delangue, Moi, Cistac,
  Rault, Louf, Funtowicz et~al.}]{wolf2019huggingface}
Thomas Wolf, Lysandre Debut, Victor Sanh, Julien Chaumond, Clement Delangue,
  Anthony Moi, Pierric Cistac, Tim Rault, R{\'e}mi Louf, Morgan Funtowicz,
  et~al. 2019.
\newblock Huggingface's transformers: State-of-the-art natural language
  processing.
\newblock \emph{arXiv preprint arXiv:1910.03771}.

\end{thebibliography}

\begin{appendices}

\section{Code details}

We release our code anonymously at \url{https://github.com/ayushk4/character-probing-pytorch} under MIT License.

The models weights, data and other dependencies required for experiment are at \url{https://github.com/ayushk4/character-probing-pytorch/releases}.

The intended use of our code is for academic research.
We consider probing publicly available PLMs, which are made available for research as well as end use cases, to be within the intended use of PLMs.

\section{Probing for Character Information} \label{appendix:Expt1}

We use off-the-shelf APIs for lemmatization and WordNet from NLTK \citep[Apache License 2.0; ][]{bird2009natural}. Our implementation uses PyTorch \citep[BSD License;][]{pytorch_paper}, HuggingFace \citep[Apache License 2.0;][]{wolf2019huggingface}  and custom APIs for GPT-J's embedding.

The probes for each MLP are trained separately starting with random initialization weights. 
We train the probe via a binary classification task via backpropagation, using the Adam optimizer \citep{kingma} with betas of 0.9 \& 0.999 and epsilon of 1e-08 without weight decay, over the standard Binary Cross Entropy loss across the predicted logits $\hat{y}_i$ and ground truth logits $y_i$.

\subsection{PLMs considered} \label{appendix:expt1_plm_details}
Details of the PLMs used along with their model-card on Huggingface:
\begin{itemize}
    \item \textbf{GPT-J:} We used the standard GPT-J with 6 Billion parameters and its reversible Byte-Pair encoding based subword tokenizer. We extracted the embeddings and have released it separately. Model Card: `EleutherAI/gpt-j-6B' under Apache 2.0 License.
    \item \textbf{GPT-2:} We consider the base model for GPT-2 with 124 Million parameters. The tokenizer used in this model is the exact same as the one used in GPT-3 and is also a subword tokenizer based on reversible Byte-Pair encoding.  Model Card: `gpt2' under Modified MIT License.
    \item \textbf{RoBERTa:} We again use the Base model for fairer comparison to the GPT-2 model with 125 Million parameters. This model has partially reversible Byte-Pair Encoding based on GPT-2's byte-pair tokenizer but with additional tokens for a BERT-like MLM discriminative pre-training. Model Card: `roberta-base' under MIT License
    \item \textbf{BERT:} The BERT-base models have roughly 110 Million parameters. 
    Both the Uncased and Cased versions of this model are considered with their Word-Piece tokenizers. For this tokenizer, we also consider the character `\#\#' while filtering out vocabulary, as it denotes the token continues on the preceding word. Model Card: `bert-base-uncased', `bert-base-cased' under Apache 2.0 License
    
\end{itemize}

\begin{table}
 \footnotesize
    \centering
    \begin{tabular}{| c | c | c |}
    \hline
     & \multicolumn{2}{|c|}{Case-Sensitive} \\
    \cline{2-3}
    Model type & PLM & Control  \\
    \hline

    GPT-J       & 94.35 & 52.76 \\
    GPT-2       & 84.69 & 51.05 \\
    RoBERTa     & 83.87 & 49.00 \\
    BERT-Cased  & 78.47 & 45.35 \\
    BERT-Uncased& 77.48 & 49.37 \\
    GloVe 300D  & 69.40 & 49.40 \\
    GloVe 100D  & 61.56 & 49.55 \\
    LXMERT      & 60.30 & 49.61 \\
    \hline
    \end{tabular}

    \caption{\label{character_results_mean_case_sensitive}
    {Results for the main probing experiment, across models.}
    }
\end{table}

\begin{table}
 \footnotesize
    \centering
    \begin{tabular}{| c | c | c | c | c |}
    \hline
     & \multicolumn{2}{|c|}{Case-insensitive} & \multicolumn{2}{|c|}{Case-Insensitive} \\
    \cline{2-3}\cline{4-5}
    Model & PLM & Control & PLM & Control \\
    \hline
    GPT-J & 0.83 & 3.12 & 1.39 & 2.27 \\
    GPT-2 & 2.01 & 3.09 & 2.21 & 2.75 \\
    RoBERTa & 2.27 & 3.13 & 2.79 & 2.46 \\
    BERT-Cased & 2.93 & 7.46 & 2.77 & 5.67 \\
    BERT-Uncased & 3.32 & 4.33 & 3.32 & 4.33 \\
    \hline
    \end{tabular}

    \caption{\label{character_results_variance}
    {Standard deviation in our probing Experiment 1, for the key models considered.}
    }
\end{table}

\begin{table}
 \footnotesize
    \centering
    \begin{tabular}{| p{20mm} | p{47mm} |}
    \hline
    Property & Statistics \\ \hline
    Dataset & Tokenizer's Vocab for each model \\
    Data-filtered & Tokens having only letters (a-z,A-Z) \\
     & \textit{GPTs, RoBERTa}: Allow preceding \.{G}\\
     & \textit{BERT}: Allow preceding `\#\#'\\
    Train-Test split & 80-20 \\
    Preprocessing & None \\
    Output labels & 26 tasks (each with binary label) \\
    Link & Model Card \& links in \S\ref{appendix:expt1_plm_details}\\
    \hline
    \end{tabular}

    \caption{\label{table:experiment1_dataset_checklist}
    {Dataset Checklist for Experiment 1.}
    }
\end{table}

\begin{table*}
 \footnotesize
    \centering
    \begin{tabular}{| c | c | c | c | c | c | c | c | c |}
    \hline
     & \multicolumn{4}{|c|}{Case-insensitive} & \multicolumn{4}{|c|}{Case-Sensitive} \\
    \cline{2-5}\cline{6-9}
    Model & \multicolumn{2}{|c|}{Lemma} & \multicolumn{2}{|c|}{Control} & \multicolumn{2}{|c|}{Lemma} & \multicolumn{2}{|c|}{Control} \\
    \cline{2-3}\cline{4-5}\cline{6-7}\cline{8-9}
    Probe & LR & \# Params & LR & \# Params & LR & \# Params & LR & \# Params \\
    \hline
    GPT-J     & 1e-4 & 240M (206M) & 1e-4 & 443M (410M) & 1e-4 & 240M (206M) & 3e-4 & 443M (410M) \\
    GPT2         & 3e-4 & 40M (39M)   & 1e-4 & 443M (410M) & 3e-4 & 40M (39M)   & 3e-4 & 443M (410M) \\
    RoBERTa      & 3e-4 & 40M (39M)   & 1e-4 & 443M (410M) & 1e-3 & 40M (39M)   & 1e-2 & 443M (410M) \\
    BERT-cased   & 1e-3 & 23M (22M)   & 3e-3 & 443M (410M) & 1e-3 & 23M (22M)   & 5e-5 & 443M (410M) \\
    BERT-uncased & 3e-3 & 25M (23M)   & 3e-4 & 443M (410M) & 3e-4 & 25M (23M)   & 1e-4 & 443M (410M) \\
    LXMERT & 1e-4 & 24M (23M)   & 3e-4 & 443M (410M) & 3e-4 & 24M (23M)   & 1e-4 & 443M (410M) \\
    GloVe 100D   & 1e-4 & 4.02M (4.00M) & 3e-4 & 12.2M (12.0M) & 3e-4 & 4.02M (4.00M) & 3e-4 & 12.2M (12.0M) \\
    GloVe 300D   & 3e-4 & 12.2M (12.0M) & 1e-4 & 12.2M (12.0M) & 3e-4 & 12.2M (12.0M) & 3e-5 & 12.2M (12.0M) \\

    \hline
    \end{tabular}

    \caption{\label{expt1_hyperparams}
    {Experiment 1 hyperparameters.}
    }
\end{table*}

\begin{itemize}
    \item \textbf{GloVe:} We experiment with the 100 and 300 dim version of 400K-Vocab GloVe trained on 6B tokens. We consider the 40k most frequent tokens in GloVe, comparable to the vocabulary sizes of the other models. GloVe version used: `Wikipedia 2014 + Gigaword 5 (6B tokens, 400K vocab, uncased, 50d, 100d, 200d, \& 300d vectors, 822 MB download): glove.6B.zip' \footnote{Accessible at nlp.stanford.edu/projects/glove/, Apache v2.0 License}
    \item \textbf{LXMERT:} We use the uncased version of LXMERT-base model and, as with the BERT model, filter out `\#\#' preceding symbols. Model Card: `unc-nlp/lxmert-base-uncased' under 
\end{itemize}

\begin{table}
 \footnotesize
    \centering
    \begin{tabular}{| p{20mm} | p{47mm} |}
    \hline
    Property & Statistics \\ \hline
    Train Sentences & 14986 \\
    Train Tokens & 219553 \\
    Valid Sentences & 3465 \\
    Valid Tokens & 55043 \\
    Test Sentences & 3683 \\
    Test Tokens & 50349 \\
    NER Tags & 5 \\
    PoS Tags & 45 \\
    Preprocessing & None \\
    Link & github: davidsbatista/NER-datasets \\
    \hline
    \end{tabular}

    \caption{\label{conll_2003_statistics}
    {Dataset Checklist for training POS/NER CoNLL set.}
    }
\end{table}

\begin{figure}[t]
    \centering
    \includegraphics[width=0.9\columnwidth]{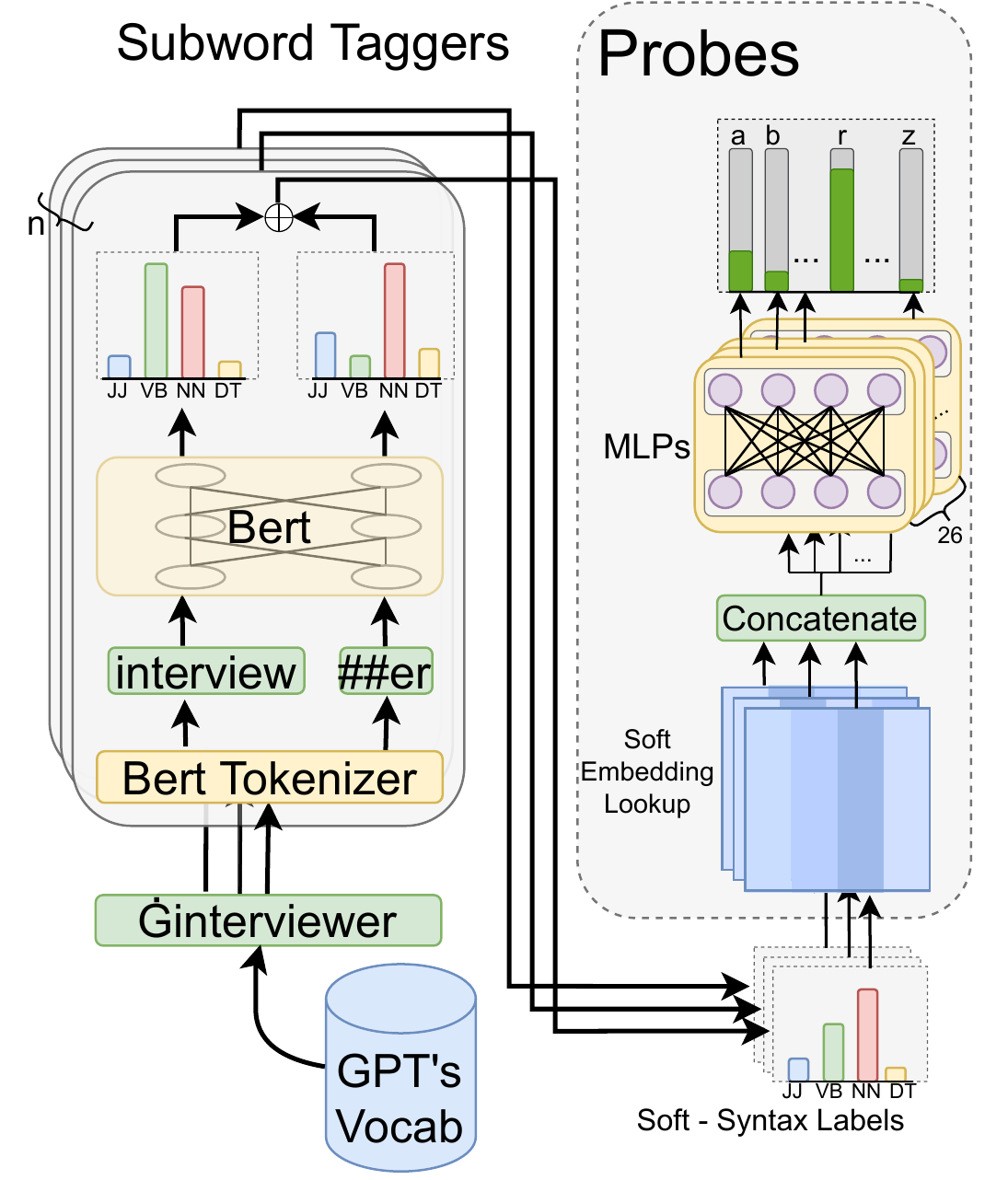}
    \caption{Experiment 2: syntax baselines with BERT-sentence and BERT-token custom taggers.}
    \label{fig:expt_2_diagram}
\end{figure}

\subsection{Hyperparameter and other Details}\label{appendix_B2_Hyperparams_expt1}

Each probe is trained for 5 epochs, with 128 batch-size. The Learning rate is tuned over averaged Macro-F1 in the grid $\{1e-5, 3e-5, 5e-5, 1e-4, 3e-4, 1e-3, 3e-3, 1e-2, 3e-2\}$. We trained the probe on the best hyperparameter settings across 5 different train-test splits and seeds. Table \ref{expt1_hyperparams} shows the best learning rates and the number of parameters (and frozen-parameters) in the probe. For all the control embedding, we assume the same dimension as the largest model (4096) and considered a maximum vocab of 100k, even though only the first few thousand might be used. 
These experiments take less than 20 minutes for each run and require less than 12 GB of GPU memory.
They were run on a mix of NVidia Tesla K80, GTX 1080 Ti, P100, V100 GPUs with Dell R740 and Intel Xeon CPUs.

Table \ref{character_results_mean_case_sensitive} shows the result of the probe in a case-sensitive setting. The case-insensitive probe treats both "Cat" and "cat" both as a hit for "c". The case-sensitive probe treats only "cat" (not "Cat") as a hit for "c". Note that performance is the same for BERT-Uncased since it does not distinguish between these conditions.

\section{Multilingual Analyses}\label{appendix_Multilingual}

\paragraph{Model Details:} We only consider mBART \citep{liu-etal-2020-multilingual-denoising} with 610M parameters and 250k vocab size. Its model card in Huggingface is `facebook/mbart-large-cc25', without any mention of its license. Its tokenizer is a reversible one, similar to GPT, except that it encodes preceding space with `$\_$'.

\paragraph{Languages:} For the non-Latin scripts considered, we only consider those characters with more than 250 occurrences in the tokenizer's vocabulary. We consider the experiment case-insensitive (by lowercasing the string) across scripts that have lowercase and uppercase characters.

\paragraph{Hyperparameters:} 
Each probe is trained for 5 epochs, with 128 batch-size. The learning rate is tuned over averaged Macro-F1 in the grid $\{1e-5, 3e-5, 5e-5, 1e-4, 3e-4, 1e-3, 3e-3, 1e-2, 3e-2\}$. We trained the probe on the best hyperparameter settings across 5 different train-test splits and seeds. Table \ref{expt3_hyperparams} shows these best learning rates and the number of parameters (and frozen parameters) in the probe. For all the control embedding, we assume the same dimension as the largest model (1024) and considered a maximum vocab of 300k, even though only a few thousand are used.
These experiments take less than 20 minutes for each run requiring less than 12 GB of GPU memory and were run on a mix of NVidia Tesla K80, GTX 1080 Ti, P100, V100 GPUs with Dell R740 and Intel Xeon CPUs.

\begin{table}
 \footnotesize
    \centering
    \begin{tabular}{| c | c | c | c |}
    \hline
    Script & PLM & Control \\
    \hline
    Latin (English chars)      & 3.28 & 7.21 \\
     Devanagari & 6.58 & 5.43 \\
     Arabic     & 10.50 & 2.99 \\
     Cyrillic   & 3.79 & 5.31 \\
    \hline
    \end{tabular}
    \caption{\label{multilingual_variability}
    {Standard Deviation for Multilingual BART experiment.}
    }
\end{table}

\begin{table*}
 \footnotesize
    \centering
    \begin{tabular}{| c | c | c | c | c |}
    \hline
     & \multicolumn{2}{|c|}{PLM} & \multicolumn{2}{|c|}{Control}\\
    \cline{2-3}\cline{4-5}
     Script     & LR & \# Params & LR & \# Params\\
    \hline
    Latin      & 3e-4 & 258M (256M) & 1e-2 & 309M (307M) \\
     Devanagari & 3e-4 & 258M (256M) & 1e-3 & 309M (307M) \\
     Arabic     & 3e-4 & 258M (256M) & 3e-3 & 309M (307M) \\
     Cyrillic   & 3e-4 & 258M (256M) & 3e-4 & 309M (307M) \\

    \hline
    \end{tabular}

    \caption{\label{expt3_hyperparams}
    {Multilingual Hyperparameters. Number of parameters (with frozen parameters in parenthesis) is denoted per probe. }
    }
\end{table*}

\section{Syntax Baseline for Character information} \label{appendix:Expt2}

\begin{table*}
 \footnotesize
    \centering
    \begin{tabular}{| c | c | c | c | c | c | c | c |}
    \hline
    
    \textbf{Model Type} & \# Epochs & Batch Size & LR & Dev $F1_{Wtd}$ & Dev $F1_{Macro}$ & Test $F1_{Wtd}$ & Test $F1_{Macro}$ \\
    \hline
    BERT-sentence (PoS) & 10 & 32 & 1e-5 & 98.17 & 94.80 & 93.42 & 87.40 \\
    BERT-token (PoS) & 10 & 32 & 1e-5 & 76.42 & 56.75 & 77.24 & 56.74 \\
    GPT-J MLP  (PoS) & 20 & 64 & 1e-4 & 62.90 & 68.72  & 60.15 & 69.14 \\
    \hline
    BERT-sentence (NER) & 10 & 32 & 1e-5 & 97.88 & 93.18 & 96.02 & 86.92 \\
    BERT-token (NER) & 10 & 32 & 1e-5 & 83.50 & 56.97 & 81.57 & 54.88 \\
    GPT-J MLP  (NER) & 20 & 64 & 5e-5 & 85.59 & 63.56 & 82.71 & 57.34 \\
    \hline
    \end{tabular}

    \caption{\label{custom_model_performance}
    {Labels from POS/NER labels. LR denotes learning rate}
    }
\end{table*}

\begin{table*}
 \footnotesize
    \centering
    \begin{tabular}{| c | c | c | c | c | c |}
    \hline
    \textbf{Split Type} & SpaCy & BERT-sentence & BERT-token & GPT-J & Control \\
    \hline
    \hline
    \multicolumn{6}{|c|}{Aggregate across 26 characters} \\
    \hline
    F1 & 52.338 & 55.008 & 59.7525 & \textbf{61.2395} & 49.6772 \\
    \hline
    \hline
    \multicolumn{6}{|c|}{Best performing ones} \\
    \hline
    s & 64.5967 & 60.7179 & \textbf{70.3299} & 66.8159 & 40.3154 \\
    y & 61.9632 & 60.3871 & \textbf{67.1591} & 64.8863 & 48.6838 \\
    e & 62.0518 & 57.7531 & \textbf{64.6152} & 62.3213 & 47.2712 \\
    t & 60.6848 & 54.3826 & \textbf{64.0681} & 60.7345 & 48.4873 \\
    p & 50.235 & 55.2361 & \textbf{63.9658} & 60.5067 & 46.5612 \\
    i & 60.8024 & 56.4055 & \textbf{63.3518} & 61.6032 & 42.8155 \\
    \hline
    \hline
    \multicolumn{6}{|c|}{Worst performing ones} \\
    \hline
    w & 45.748 & 52.7235 & 57.6919 & \textbf{58.2666} & 48.6947 \\
    q & 43.7924 & 56.5274 & \textbf{57.5407} & 53.5437 & 49.2841 \\
    k & 47.7873 & 49.3832 & \textbf{57.3084} & 55.9559 & 46.2371 \\
    o & 52.9403 & 53.6138 & \textbf{56.8312} & 55.6293 & 43.5871 \\
    b & 48.9159 & \textbf{56.739} & 56.3873 & 55.1265 & 48.252 \\
    m & 48.1349 & 53.4036 & \textbf{56.2846} & 55.6094 & 46.1084 \\
    \hline
    \end{tabular}

    \caption{\label{appendix:pos_ner_results}
    {Syntax baseline: Probing over syntax label distribution.}
    }
\end{table*}

\subsection{Custom syntax taggers}

First we consider an off-the-shelf SpaCy model with 3 features for each token: NER, PoS, and Coarse-Grained PoS tags. 
Before running this model, we remove the preceding whitespace characters in the token, if present. 
The resultant features are discrete one-hot feature vectors over labels.
The SpaCy tagger is not perfectly suited to our task since it operates at the word level, whereas we are concerned with obtaining a subword token's embeddings.
To solve that problem, we also built 3 custom taggers for obtaining PoS and NER labels on subword tokens.
These taggers take (a subword) token's model embedding as input and output a vector of probabilities over part of speech and named entity categories.

To build our custom GPT-J-Tagger, we train an MLP to predict PoS and NER label based on GPT-J's static embedding layer for each token.
The tagger is trained on the CoNLL 2003 dataset's train and evaluation splits \citep{sang2003introduction}, which contains part of speech and named entity information. Unlike the SpaCy tagger, our custom GPT-J-Tagger outputs a probability distribution over categories so we can use this distribution over labels as the vector of interest, rather than a one-hot vector.

Table \ref{custom_model_performance} show the performance of the tagger's performance \textit{qua} tagger. Table \ref{conll_2003_statistics} shows the Dataset Checklist for this experiment.
To build the BERT sequence-labeling tagger, we fine-tuned a BERT sequence labeling model for the PoS and NER tasks, in order to output a label for each (sub-word) token in a sentence. 
When extracting syntactic features for this model, we first do the same pre-processing of removing the special preceding whitespace of GPT's tokens as SpaCy before input into the BERT model. 
Since BERT's tokenizer could have more than one token for a single GPT-J's token, we consider the average of the logits as the pre-softmaxed feature vector.

In addition to the BERT sentence-level tagger, we consider a BERT token classifier model fine-tuned for NER and PoS at token level rather than at sentence level. This token-level model does not leverage context to deduce the label, and is closer to how we use this model later to get features for predicting NER/PoS features.

\begin{table*}
 \footnotesize
    \centering
    \begin{tabular}{| c | c | c | c | c | c |}
    \hline
    \textbf{Split Type} & SpaCy & BERT-sentence & BERT-token & GPT-J & Control \\
    \hline
    \hline
    \multicolumn{6}{|c|}{Aggregate across 26 letters} \\
    \hline
    F1 & 4.4354 & 2.9588 & 3.7989 & 2.724 & 4.3973 \\
    \hline
    \hline
    \multicolumn{6}{|c|}{Best performing ones} \\
    \hline
    s & 0.6947 & 1.2941 & 0.4853 & 0.6514 & 5.5055 \\
    y & 1.8665 & 1.6406 & 0.5697 & 1.4251 & 3.2417 \\
    e & 0.6645 & 0.8544 & 0.3245 & 0.3233 & 1.8349 \\
    t & 0.2643 & 3.4695 & 0.9129 & 0.5924 & 1.7645 \\
    p & 6.1928 & 1.1628 & 0.5669 & 0.2985 & 3.7013 \\
    i & 0.512 & 1.4392 & 0.5998 & 0.4867 & 5.5685 \\
    \hline
    \hline
    \multicolumn{6}{|c|}{Worst performing ones} \\
    \hline
    w & 4.9794 & 2.2996 & 1.9614 & 1.9536 & 1.7453 \\
    q & 2.7071 & 3.4438 & 4.5954 & 4.7932 & 5.5068 \\
    k & 2.9332 & 6.885 & 2.0885 & 1.6864 & 1.6311 \\
    o & 6.24 & 1.6009 & 1.0449 & 0.463 & 3.5961 \\
    b & 4.0455 & 1.5597 & 1.4074 & 2.0701 & 2.7857 \\
    m & 7.2995 & 2.4854 & 2.1762 & 1.0948 & 6.152 \\
    \hline
    \end{tabular}

    \caption{\label{appendix:pos_ner_variance}
    {Standard Deviation of POS/NER labels}
    }
\end{table*}

\subsection{Results and Hyperparameters}

We use off-the-shelf APIs for lemmatization and WordNet from NLTK. Our implementation uses PyTorch \cite{pytorch_paper}, HuggingFace \citep{wolf2019huggingface} and custom APIs (now released) for GPT-J's embedding. The hyperparameter tuning was done on the dev set for only the learning rate in the grid $\{1e-5, 3e-5, 1e-4\}$ for BERT and $\{1e-5, 3e-5, 5e-5, 1e-4, 3e-4, 1e-3, 3e-3, 1e-2, 3e-2\}$ for GPT-J. Our MLP model is 3-layered with SELU and Tanh activation and 0.1 Dropout before the last layer. Our BERT-Model is initialized with `bert-base-cased` from Huggingface with default values of hyperparameters. Our implementation was done using PyTorch and optimized via Adam with betas of 0.9 \& 0.999 and epsilon of 1e-08 without weight decay over the standard Cross Entropy loss. We set the batch size to 32 sentences for BERT and 64 for GPT-J. All the experiments can be done within 16GB of GPU memory and no run individually takes more than 2 hours. We release these models along with our codebase with instructions to run them. 

Table \ref{custom_model_performance} shows the performance of these NER and PoS models. As expected, the BERT-sentence model performs the best on both the tasks as it leverages the context while tagging. GPT-J slightly outperfoms BERT-token on both the tasks. Note that these performances are not comparable as their tokenizations differ and only one of the models leverages context to predict NER and PoS tags.

\subsection{Method}

Assume we have $m$ syntactic features. Consider the tokenizer Vocabulary $V$ (with only alphabetic tokens) and the $D_{\alpha}$ datapoint pairs for each letter $\alpha$ of the lowercased English alphabet. For each token-label pair $(w_i, y_i)$, we obtain the $m$ syntactic features of the word $\{x^{(1)}_i, x^{(2)}_i \dots x^{(m)}_i\}$ using the trained models to tag the features.

We train a classifier to predict whether a character $\alpha$ is present in the token $w_i$ using only its syntactic features. Assume randomly initialized `trainable' embeddings $\{E_1, E_2 \dots E_m\}$ for each of the $m$ syntactic features. We predict the logits $\hat{y}_i$ for token $w_i$ over each letter $\alpha$ using an MLP classifier over the embeddings:

$$\hat{y}_i = \sigma(MLP_{\alpha}([E_1^T x^{(1)}_i\ ;\ \dots\ ;\ E_m^T x^{(m)}_i\ ]))$$

 Each syntactic feature $x^{(j)}_i$ is a vector denoting probability distribution of a token over the corresponding feature labels (including being a one-hot vector), this is crucial because a token (especially subword-token) might have different labels depending on the context.

We train different MLPs and Embeddings from scratch for each alphabet $\alpha$ with no shared parameters across the (case-insensitive) 26 English characters. We train our model for binary classification via backpropagation over the standard Binary Cross Entropy loss across the predicted logits $\hat{y}_i$ and ground truth logits $y_i$.

As before, for each character we create a balanced dataset consisting of an equal number of positive and negative examples, where each example is made up entirely of either English characters or whitespace. 
These are randomly divided into training and test split sucht that we keep words with with the same lemmas in the same split. As a control task, we randomly assign the syntactic features for each token.
We set the batch size for runs with one-hot vectors as features to 128 and to 64 for others, the learning rate is tuned in $\{1e-5, 3e-5, 1e-4, 3e-4, 1e-3, 3e-3, 1e-2\}$ for all the features over the metric of Averaged F1-Scores across the 26 English letters. The best learning rates for SpaCy, BERT-sentence, BERT-token, GPT-J and Control were found to be 1e-3, 1e-3, 3e-3, 1e-4, 1e-2, respectively. Using Adam Optimizer we train each of the 26 models for 5 epochs with betas of 0.9 \& 0.999 and epsilon of 1e-8. Our implementation is done using PyTorch and Huggingface. Finally for the best hyperparameter, we perform 5 runs with different train/test splits and seeds. Our MLP model is 3-layered with SELU and Tanh activation and 0.1 Dropout before the last layer. 

Tables \ref{appendix:pos_ner_results} and \ref{appendix:pos_ner_variance} show the mean and variance of the results over the 4 taggers and control task. We also show the performance over the best-performing and worst-performing characters.

\section{Variability of Tokenization} \label{appendix:Expt3}

\subsection{Quantifying variability in the Pile Corpus}

To quantify the variability in the tokenization of frequent words in corpora comparable to the corpora used to train these models,
we consider 1/6th of the publicly available Pile Corpus used to train GPT-J (~250 GB of text).
For our analysis we consider 500 frequent words of 8+ characters (as measured using Google Ngrams) since long words are more likely to be the source of variability.

For each target word, we first case-insensitively detect each of its occurrences in the sub-corpus.
In order to also account for spelling errors, we used case-insensitive fuzzy search, allowing matches for substrings up to 1 Levenshtein distance away.
Over these occurrences, we discard those where the substring is part of a bigger word, such as `differentiation' for the target word `different' or if the fuzzy match has whitespaces.
 
Once we have such occurrences, we want to obtain the tokenization of the target word in the context.
For each word in the set of matches, if the matched substring ends with a non-valid character for our probing task, we delete the final character.
This allows for matches of [somethin', somethin", somethin] all to be considered as the string `somethin'. 
We also account for the factors that leads to differing tokenization, such as preceding whitespaces.

Now, for each of the target words, we have a list of probable tokenization at most 1 Levenshtein distance away.
Since two target words such as `projection' and `protection' could themselves be at 1 Levenshtein distance, these may act as what we call ``pseudo matches" for each other. So we consider only one of these two from our target list, leading to 466 word down from 500 words. Now, for each of these target words, we count the number of possible unique tokenizations.

For each of these 466 target words, we also obtain a list of words from WordNet, which are 1 Levenshtein distance away. We treat this word list as the pseudo-match list. We also consider the number of tokenizations for each target word, excluding their pseudo-match list as well as by excluding all those which are equally close to or closer to a word in the pseudo-match list than they are to the target word. 
We also compute the statistics of those with exact matches.

Table \ref{tokenization_variance_statistics} shows these statistics for the target words. On average, a target word is expected to have 213 different tokenizations depending on the context. We observe that, while one may expect the number of tokenizations to go up with the number of characters in the target word, it doesn't perfectly increase monotonically. This is because the number of occurrences of the target word dictates the number of tokenization it will have.
Unsurprisingly, we see a consistent trend that the number of tokenization greatly increases with increasing occurrences. 

We observe three factors contributing to a remarkably large number of tokenizations. First, Case-Sensitive tokenization leads to up to 6 different tokenizations for each of the target words. 
Second, context-dependent tokenization increases the expected number of different tokenizations to 12.91.
The rest of the tokenizations are likely due to misspellings or variants.

Our analyses were sped up using multiprocessing and fuzzy regex.
To do so, we split the sub-corpus across multiple pieces. These runs take about 3 days across 40 CPU Cores, 60 GB of RAM and less than 600GB hard disk space. 
We report the mean and standard deviation for the number of tokenizations a word has across the portion of the Pile corpus considered. 
These are also reported as a function of word length and its frequency of occurrence in the corpus.

Tables \ref{tokenization_variance_statistics} and \ref{tokenization_variance_variability} shows these scores. The `All matches' field considers the unique tokenizations of all matched substrings including those at 1 (case and whitespace insensitive) Levenshtein distance away. 
These word at 1 Levenshtein distance could be either misspellings or a different English word (for example an occurrence of the word `projection' for target word `protection'). 
The latter of these are identified using the Wordnet dictionary and the statistics recalculated and shown in the column `Matches except pseudo'. 
Some of the misspellings contributing to this score could be misspellinsg of either the target word or of one of the other English words at 1 Levenshtein distance away (`prohection' could be a misspelling of either `projection' or `protection' being at distance 1 from both). 
Such occurrences are removed, with statistics recomputed for the column `Matches closer pseudo'. 
The column `Exact contain' considers only those occurrences, which contain the exact target word (case-insensitively) in the string ignoring whitespaces. The `Exact match' column does not consider occurrences involving a preceding whitespace.

Table \ref{tokenization_variance_examples} shows some examples of variation in tokenization.

\newpage

\subsection{Algorithm for increasing tokenization variability}

\begin{algorithm}
\scriptsize
\caption{A simplified version of subword Tokenization with controllable variability}\label{alg:controllable_tokenization}
\begin{algorithmic}
\Require $0<=\rho <= 1$
\Procedure{YourFunction}{$sentence$}
\State $tokens \gets List()$
\State $words \gets wordTokenize(sentences)$
\For{each $w$ in $words$}
    \State $u \sim Uniform[0,1]$
    \If{$u < \rho$}
        \State $V \gets GPTJ.Vocab$
        \State $filter(V, \lambda x. isAlphabetic(x))$
        \State $Choices \gets List()$
        \For{$i$ in $1, 2 \dots (w.length()-1)$}
            \If{$w[$:$i] \in V$ \& $w[i$:$] \in V$}
                \State $push(Choices, w[$:$i], w[i$:$])$
            \EndIf
        \EndFor
        \If{$\lnot isEmpty(Choices)$}
            \State $s \sim Choices$
            \State $tokens \gets Merge(tokens, s)$
            \State $continue$
        \EndIf
    \EndIf
    \State $s \gets GPTJ.Tokenize(w)$
    \State $tokens \gets Merge(tokens, s)$
\EndFor

\EndProcedure
\end{algorithmic}
\end{algorithm}

\begin{table*}
 \footnotesize
    \centering
    \begin{tabular}{| c | p{1.5cm} | p{2cm} | p{1.5cm} | p{1.5cm} | p{1cm} | p{1cm} |}    \hline
    \textbf{Measure} & All Matches & Matches Excluding Pseudo & Matches Closer Pseudo & Exact Contain & Exact Match & Num Words\\
    \hline
    Aggregate       & 232.90 & 229.70 & 213.74 & 17.91 & 5.97 & 466 \\
    \hline
    7 Length words  & 297.50 & 271.00 & 223.50 & 22.00 & 6.5 & 2 \\
    8 Length words  & 332.29 & 325.68 & 288.07 & 25.00 & 7.89 & 28 \\
    9 Length words  & 231.48 & 227.78 & 206.95 & 16.94 & 5.93 & 190 \\
    10 Length words & 225.51 & 222.58 & 209.53 & 17.97 & 5.87 & 127 \\
    11 Length words & 213.28 & 211.02 & 202.97 & 17.88 & 5.85 & 61 \\
    12 Length words & 224.14 & 223.54 & 218.64 & 18.25 & 5.79 & 28 \\
    13 Length words & 218.14 & 217.00 & 214.76 & 16.57 & 5.19 & 21 \\
    14 Length words & 238.33 & 238.33 & 238.33 & 16.67 & 5.00 & 9 \\
    \hline
    exp(12) occurrence & 88.70  & 86.67  & 82.11  & 10.33 & 5.90 & 27 \\
    exp(13) occurrence & 155.78 & 153.87 & 146.55 & 13.61 & 5.15 & 74 \\
    exp(14) occurrence & 210.36 & 207.51 & 195.74 & 16.70 & 5.75 & 174 \\
    exp(15) occurrence & 278.88 & 275.00 & 251.69 & 19.91 & 5.96 & 139 \\
    exp(16) occurrence & 370.02 & 365.04 & 336.48 & 26.62 & 8.56 & 52 \\
    \hline
    \end{tabular}

    \caption{\label{tokenization_variance_statistics}
    {Tokenization variance statistics - mean score.}
    }
\end{table*}

\begin{table*}
 \footnotesize
    \centering
    \begin{tabular}{| c | p{2cm} | p{2cm} | p{2cm} | p{2cm} | p{2cm} |}
    \hline
    \textbf{Measure} & All Matches & Matches Excluding Pseudo & Matches Closer Pseudo & Exact Contain & Exact Match \\
    \hline
    Aggregate   & 95.12  & 94.29  & 91.26  & 17.91 & 2.67 \\
    \hline
    7 Length words  & 155.50 & 129.00 & 81.50  & 13.00 & 2.50 \\
    8 Length words  & 100.90 & 99.17  & 91.19  & 8.46  & 2.47\\
    9 Length words  & 90.97  & 90.00  & 86.03  & 7.34  & 2.50 \\
    10 Length words & 88.56  & 89.04  & 90.71  & 7.86  & 2.75 \\
    11 Length words & 107.55 & 107.65 & 108.46 & 8.77  & 2.84 \\
    12 Length words & 63.25  & 63.53  & 62.53  & 8.26  & 2.82 \\
    13 Length words & 81.22  & 81.30  & 82.20  & 7.82  & 2.59 \\
    14 Length words & 62.48  & 62.48  & 62.48  & 4.52  & 1.05 \\
    \hline
    exp(12) occurrence & 38.59  & 37.65  & 34.60  & 3.15 & 1.26 \\
    exp(13) occurrence & 39.75  & 39.13  & 39.36  & 4.92 & 2.10 \\
    exp(14) occurrence & 51.84  & 52.17  & 53.73  & 6.19 & 2.51 \\
    exp(15) occurrence & 70.46  & 70.59  & 77.22  & 7.86 & 2.38 \\
    exp(16) occurrence & 101.86 & 100.38 & 103.83 & 9.99 & 3.44\\
    \hline
    \end{tabular}

    \caption{\label{tokenization_variance_variability}
    {Variability across target words in tokenization variance statistics.}
    }
\end{table*}

\begin{table*}
 \footnotesize
    \centering
    \begin{tabular}{| c c | c c  |}
    \hline
    String & Tokenization & String & Tokenization  \\
    \hline
    \multicolumn{2}{|c|}{signature} & \multicolumn{2}{|c|}{playstation} \\ 
    \hline
    \multicolumn{2}{|c|}{Exact match case insensitive} & \multicolumn{2}{|c|}{Exact match case insensitive}  \\
    \hline
    "SIGNATURE" & ["SIGN", "ATURE"] & "playstation" & ["play", "station"] \\
    "sIGNATURE" & ["s", "IGN", "ATURE"] & "PLaySTATION" & ["PL", "ay", "ST", "ATION"]  \\
    "SigNature" & ["S", "ig", "Nature"] & "playStation" & ["play", "Station"]"\\
    "Signature" & ["Sign", "ature"] & "PLAYSTATION" & ["PLAY", "ST", "ATION"]  \\
    "SIgnature" & ["SI", "gn", "ature"] & "Playstation" & ["Play", "station"]\\
    "signature" & ["sign", "ature"] & "PlayStation" & ["Play", "Station"]   \\
    \hline
    \multicolumn{2}{|c|}{Exact match and whitespaces} & \multicolumn{2}{|c|}{Exact match and whitespaces} \\
    \hline
    " signature" & ["Ġsignature"] & " Playstation" & ["ĠPlaystation"] \\
    " Signature" & ["ĠSignature"] & " PLayStation" & ["ĠPL", "ay", "Station"] \\
    " SigNature" & ["ĠSig", "Nature"] & " PLAYstation" & ["ĠPLAY", "station"]  \\
    " signaTure" & ["Ġsign", "a", "T", "ure"] & " PLAYSTATION" & ["ĠPLAY", "ST", "ATION"]  \\
    " SIGNATure" & ["ĠSIGN", "AT", "ure"] & " PlayStation" & ["ĠPlayStation"] \\
    " SiGNATURE" & ["ĠSi", "GN", "ATURE"] & " plAYsTaTion" & ["Ġpl", "AY", "s", "Ta", "T", "ion"]  \\
    " SIGNATURE" & ["ĠSIGN", "ATURE"] & " playStation" & ["Ġplay", "Station"] \\
    " signAture" & ["Ġsign", "At", "ure"] & " playstation" & ["Ġplay", "station"] \\
    " SIGNature" & ["ĠSIGN", "ature"] & " PLaystation" & ["ĠPL", "ay", "station"] \\
    " sIgnature" & ["Ġs", "Ign", "ature"] & " PlaySTation" & ["ĠPlay", "ST", "ation"] \\
    \hline
    \multicolumn{2}{|c|}{Fuzzy match and misspellings} & \multicolumn{2}{|c|}{Fuzzy match and misspellings} \\
    \hline
    "S1GNATURE" & ["S", "1", "GN", "ATURE"] & "Play-station" & ["Play", "-", "station"] \\
    " SIGNATUTRE" & ["ĠSIGN", "AT", "UT", "RE"] & " PLAY-STATION" & ["ĠPLAY", "-", "ST", "ATION"] \\
    " signatyure" & ["Ġsign", "at", "y", "ure"] & "play-station" & ["play", "-", "station"]  \\ 
    " signatre" & ["Ġsign", "atre"] & " Play-station" & ["ĠPlay", "-", "station"] \\
    "Signiature" & ["Sign", "i", "ature"] & " play-station" & ["Ġplay", "-", "station"]  \\
    " signnature" & ["Ġsign", "nature"] & "Play-Station" & ["Play", "-", "Station"]  \\
    " signatrre" & ["Ġsign", "at", "r", "re"] & "Play]station" & ["Play", "]", "station"] \\
    " sigature" & ["Ġsig", "ature"] & " Playst4tion" & ["ĠPlay", "st", "4", "tion"] \\
    " Sign(ature" & ["ĠSign", "(", "ature"] & " PlayStati0n" & ["ĠPlay", "St", "ati", "0", "n"]  \\
    "signnature" & ["sign", "nature"] & " Play-Station" & ["ĠPlay", "-", "Station"] \\
    "SIG(NATURE" & ["S", "IG", "(", "NAT", "URE"] & "Playstaton" & ["Play", "st", "aton"]  \\
    " Si2nature" & ["ĠSi", "2", "nature"] & " play.Station" & ["Ġplay", ".", "Station"] \\
    "Singnature" & ["Sing", "nature"] & " playstaton" & ["Ġplay", "st", "aton"] \\
    " signatuure" & ["Ġsign", "atu", "ure"] & " PLAYTSTATION" & ["ĠPLAY", "T", "ST", "ATION"]  \\
    " Signaturs" & ["ĠSign", "at", "urs"] & "playstatiom" & ["play", "st", "ati", "om"]   \\
    " sigNUTure" & ["Ġsig", "N", "UT", "ure"] & "playsstation" & ["plays", "station"] \\
    \hline
    \end{tabular}

    \caption{\label{tokenization_variance_examples}
    {Some examples of variations in tokenization for two example words.}
    }
\end{table*}

\end{appendices}

\end{document}